\documentclass{article}
\usepackage{ijcai24}

\usepackage{times}
\usepackage{soul}
\usepackage{url}
\usepackage[hidelinks]{hyperref}
\usepackage[utf8]{inputenc}
\usepackage[small]{caption}
\usepackage{graphicx}
\usepackage{amsmath}
\usepackage{amsthm}
\usepackage{booktabs}
\usepackage{algorithm}
\usepackage{algorithmic}
\usepackage[switch]{lineno}
\usepackage{natbib}

\usepackage{graphicx}
\usepackage{enumitem}

\usepackage{dsfont}
\usepackage{amsmath}
\usepackage{amsfonts}
\usepackage{amsthm}

\newtheorem{problem}{Problem}

\newcommand{\continuation}{??}

\usepackage{subcaption}
\usepackage{booktabs}
\usepackage{lipsum}
\usepackage{xcolor}

\usepackage{amsmath,amsfonts,bm}

\def\eqref#1{equation~\ref{#1}}

\def\1{\bm{1}}

\DeclareMathAlphabet{\mathsfit}{\encodingdefault}{\sfdefault}{m}{sl}
\SetMathAlphabet{\mathsfit}{bold}{\encodingdefault}{\sfdefault}{bx}{n}

\DeclareMathOperator*{\argmax}{arg\,max}
\DeclareMathOperator*{\argmin}{arg\,min}

\usepackage{tipa}
\newcommand{\joiner}[0]{{\ensuremath{\mathcal{J}}}}

\usepackage{pgf}

\title{Learning Translations: Emergent Communication Pretraining\\for Cooperative Language Acquisition}
\author {
    Dylan Cope\footnote{
        Correspondence: \url{dylan.cope@kcl.ac.uk}. DC is supported by the UKRI Centre for Doctoral Training in Safe and Trusted AI (EPSRC Project EP/S023356/1). Preprint (under review). 
    } \And
    Peter McBurney
    \affiliations {
        King's College London
    }
}

\begin{document}
\maketitle
\begin{abstract}

In Emergent Communication (EC) agents learn to communicate with one another, but the protocols that they develop are specialised to their training community. This observation led to research into Zero-Shot Coordination (ZSC) for learning communication strategies that are robust to agents not encountered during training.
However, ZSC typically assumes that no prior data is available about the agents that will be encountered in the zero-shot setting. In many cases, this presents an unnecessarily hard problem and rules out communication via preestablished conventions. We propose a novel AI challenge called a \textit{Cooperative Language Acquisition Problem} (CLAP) in which the ZSC assumptions are relaxed by allowing a `joiner' agent to learn from a dataset of interactions between agents in a target community.
We propose and compare two methods for solving CLAPs: Imitation Learning (IL), and \textit{Emergent Communication pretraining and Translation Learning} (ECTL), in which an agent is trained in self-play with EC and then learns from the data to translate between the emergent protocol and the target community's protocol.

\end{abstract}

\section{Introduction}

Creating teams of artificial agents that can communicate and cooperate has been a long-standing area of interest in multi-agent systems research.
Advances in multi-agent reinforcement learning have enabled researchers in the field of Emergent Communication (EC) to train such teams in ever more complex domains \citep{wagner_progress_2003,foerster_learning_2016, sukhbaatar_learning_2016, jaques_social_2019,lazaridou_emergent_2020}.
These agents are typically trained by allowing discrete messages to be exchanged between agents.
Programmers do not assign meaning to the messages, rather, meaning emerges via the training process as communicative conventions are developed in service of solving the task.
As such, the mapping between meanings and messages is arbitrary, and any permutation of a learned protocol is equally likely to appear across different training runs \citep{bullard_quasi-equivalence_2021}.
The result is that the learned conventions established within a training community will be very unlikely to work with new agents, and by default, the EC trained agents will be incapable of adapting.

In response to this, many researchers have become interested in devising methods in which agents learn communicative strategies that can adapt to this \textit{Zero-Shot Coordination} (ZSC) setting
\citep{li_cooperative_2023,hu_off-belief_2021, hu_other-play_2020, ossenkopf_comaze_2020, cope_learning_2020, bullard_exploring_2020}.
ZSC algorithms typically aim to successfully communicate with an unknown agent on the first encounter, without any prior information.
But in many real-world settings, this is an unnecessarily challenging assumption.
If someone is injured on a street in London, passing pedestrians can form an ad hoc team and aid the patient by speaking to each other in English to coordinate a response.
Indeed, language is arguably the most critical set of conventions that such teams can draw upon to efficiently work together.

The study of artificial agents that can form ad hoc teams is known as \textit{Ad Hoc Teamwork} (AHT) \citep{stone_ad_2010}. Similarly to ZSC, most of these algorithms aim to make as few assumptions as possible about the players that an agent may form a team with. Notably, \citet{sarratt_role_2015} applied this minimalist approach to communication.
Other work has relaxed this by assuming a prior known communication protocol \citep{barrett_communicating_2014, mirsky_penny_2020}.

In this work, we present a novel AI challenge that we call a \textit{Cooperative Language Acquisition Problem} (CLAP).
Here by `language acquisition' we mean learning the syntax and semantics of a preexisting communication system used by a community.
This class of problems is positioned between the challenges of ZSC and AHT.
In a CLAP, we are given a dataset of communication events between speakers and listeners in a \textit{target community} as they solve a problem. Our goal is to construct a \textit{joiner agent} that can communicate and cooperate with agents from this community.
This problem is also closely related to Imitation Learning (IL), however, most work in IL is confined to the single agent setting \citep{hussein_imitation_2017}. So to the best of our knowledge, this is the first attempt to pose an IL problem for multi-agent communication within a formal cooperative model.

Alongside defining CLAP, we outline two baseline solutions to this problem. The first uses a simple imitation learning method.
The second is a novel algorithm called \textit{Emergent Communication pretraining and Translation Learning} (ECTL). 
We introduce two environments and train target communities of agents that cooperate via a learned communication protocol.
Data is then gathered from these communities and then used to train joiner agents with ECTL and IL.
We demonstrate that ECTL is more robust than IL to expert demonstrations that give an incomplete picture of the underlying problem and that ECTL is significantly more effective than IL when communications data is limited.
Finally, we apply these methods to manually collected data and show that ECTL can learn to communicate with a human to cooperatively solve a task.


\section{Background}

\subsection{Decentralised POMDPs}

A \textit{Decentralised Partially-Observable Markov Decision Process} (Dec-POMDP) is a formal model of a cooperative environment defined as a tuple  $\mathcal{M} =(\mathcal{S},\mathcal{A},T,r,\boldsymbol{\Omega},O)$ \citep{oliehoek_concise_2016}, where $\mathcal{S}$ is a set of states, and $\mathcal{A} = \prod_i \mathcal{A}_{i}$ is a product of individual agent action sets. A \textit{joint action} $\mathbf{a} \in \mathcal{A}$ is a tuple of actions from each agent that is used to compute the environment's transition dynamics, defined by a probability distribution over states $T:\mathcal{S}\times\mathcal{A}\times\mathcal{S}\rightarrow[0,1]$. Team performance is defined by a cooperative reward function $r: \mathcal{S} \times \mathcal{A} \times \mathcal{S}$ over state transitions and joint actions. $\boldsymbol{\Omega} = \{\Omega_{i}\}$ is a set of observation sets, and $O:S\rightarrow\prod_i \Omega_i $ is an observation function.

Each agent $i$ follows a policy $\pi_i$ that maps an observation sequence (or a single observation if $i$ is \textit{memoryless}) to a distribution over its actions. A \textit{trajectory} for an agent $i$ is a sequence of observation-action-reward tuples $\tau_i \in \mathcal{T}_i = (\Omega_i\times\mathcal{A}_i\times\mathds{R})^*$. For a set of policies $\Pi = \{\pi_i\}$, a joint trajectory is $\boldsymbol{\tau} \in \mathcal{T} = (\boldsymbol{\Omega}\times\mathcal{A}\times\mathds{R})^*$, and we can denote the distribution of joint trajectories for this set of policies acting in the environment as $\mathcal{M}|_{\Pi}$. In this work, we will only consider finite-horizon Dec-POMDPs, so the lengths of trajectories will always be bounded.
The \textit{total reward} for this trajectory is the sum of rewards along the sequence, denoted $R(\boldsymbol{\tau})$.
The expected sum of rewards for a set of policies will be denoted $R(\Pi) = \mathds{E}_{\boldsymbol{\tau} \sim \mathcal{M}|_{\Pi}}[R(\boldsymbol{\tau})]$. 

\subsection{Emergent Communication}

Emergent communication is the study of agents that learn (or evolve) to make use of communication channels without previously established semantics. In a typical set-up, each agent's action set in the Dec-POMDP can be expressed as $\mathcal{A}_i = \mathcal{A}_i^e \times \mathcal{A}^c_i$, where $\mathcal{A}_{i}^c$ is a set of \textit{communicative actions}, and $\mathcal{A}_i^e$ is a set of \textit{environment actions}.
The communicative actions can further be written as the product of one-way communication channels from $i$ to $j \in C_i$ using a discrete \textit{message} alphabet $\Sigma$, i.e. $\mathcal{A}^c_i=\Sigma^{|C_i|}$.
These are cheap-talk channels, meaning there is no cost to communication.
This variant of a Dec-POMDP is known as a Dec-POMDP-Com \citep{goldman_decentralized_2004, goldman_communication-based_2008, oliehoek_concise_2016}.
The messages have no prior semantics as the transition function of the Dec-POMDP only depends on the environment actions ${\mathcal{A}_i^e}$, and agents are not programmed to send messages with any prescribed meaning.
Rather, the semantics emerge through training.

For this paper, we will consider memoryless agents that communicate within a centralised forward pass computing a joint action for the environment (similar to \citealt{goldman_optimizing_2003}).

\textbf{Policy Factorisation.}
We suppose that we can factor a memoryless communication agent's policy $\pi_i$ into an environment-level policy $\pi^e_i : \Omega_i \times \Sigma^{n_s} \rightarrow \mathcal{A}^e_i$, where $n_s$ is the number of agents that send messages to agent $i$, and a communication policy $\pi^c_i : \Omega_i \rightarrow \mathcal{A}^c_i$. 
Therefore, given a joint observation $\mathbf{o} = (o_1,\ldots,o_N) \in \prod_i \Omega_i$, the joint policy $\boldsymbol{\pi}(\mathbf{o})$ is computed by first producing the outgoing messages $M_{out}^i = \pi^c_i(o_i)$ from each sender agent $i$ to each receiver agent $j \in C_i$. A set of incoming messages $M^i_{in}$ is then constructed for each agent $i$ by passing the messages for $i$ through a communication channel function $\sigma_i : \mathcal{A}^c_i \rightarrow \Sigma$. The final environment actions are then given by $a_i = \pi^e_i(o_i, M^i_{in})$.




\textbf{Self-play.} Emergent communication training is often done in self-play, where parameters are shared between agents and centrally optimised \citep{hu_other-play_2020}. Communication can be hard to learn with Reinforcement Learning (RL) as sending messages cannot provide reward unless listeners already know how to use the information. This chicken-and-egg problem can be solved with centralised training by allowing gradients to backpropagate through communication channels. A Gumbel-Softmax \citep{jang_categorical_2017, maddison_concrete_2017} function is a popular choice as it facilitates backpropagation during training and can be discretised during evaluation.

However, as \citet{lowe_pitfalls_2019} have shown, measuring whether or not agents trained to communicate are doing so can be tricky for complex environments. To remedy this, they introduced definitions of \textit{positive listening} and \textit{positive signalling}.
In short, to be positive listening/signalling; the listening agent should change its actions according to the messages it receives and the signalling agent should change its messages depending on what it observes.

\subsection{Imitation Learning}

\begin{figure*}[t]
\centering
\captionsetup[subfigure]{justification=centering}
\begin{subfigure}{.4\textwidth}
    \centering
    \includegraphics[height=3.5cm]{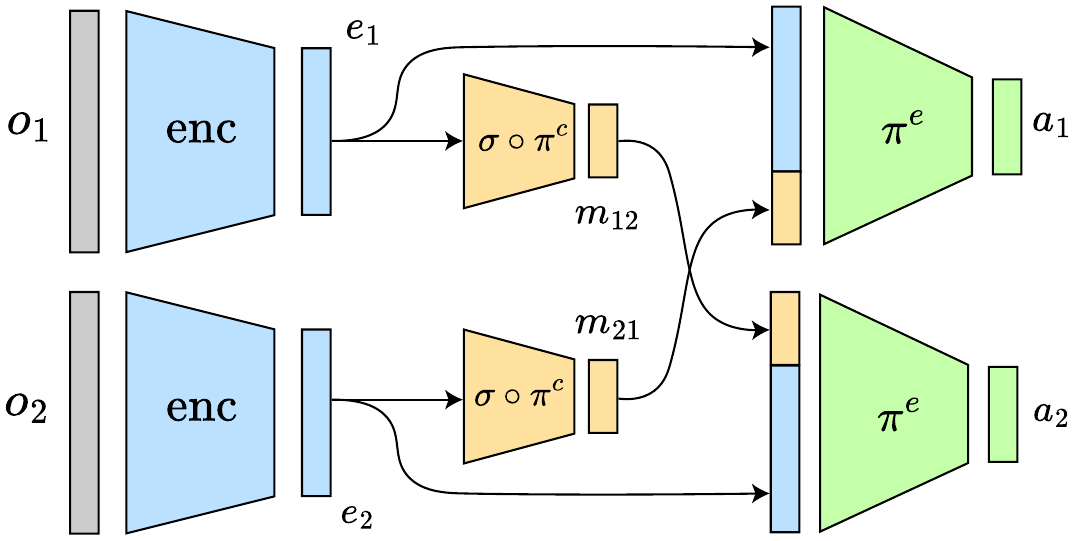}
    \caption{Joint agents architecture (two agents)\\with communication.}
    \label{fig:arch_diagram}
\end{subfigure}%
\begin{subfigure}{.3\textwidth}
    \centering
    \includegraphics[height=3.5cm]{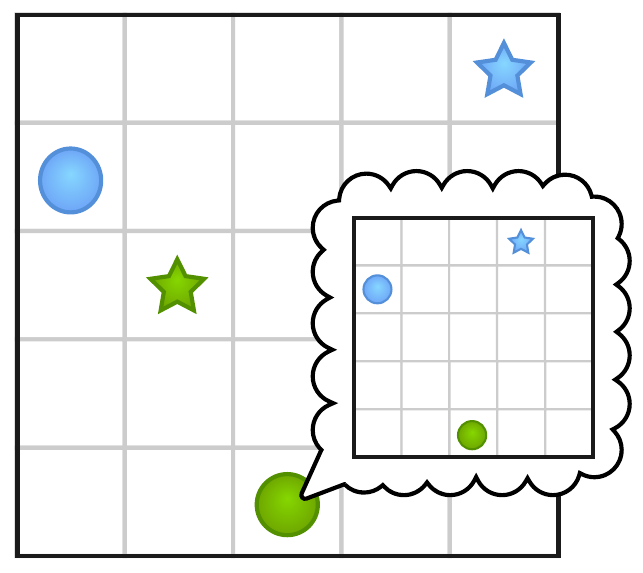}
    \caption{Illustration of the\\gridworld environment (Section \ref{sec:envs}).}
    \label{fig:toy_env_illustration}
\end{subfigure}%
\begin{subfigure}{.3\textwidth}
    \centering
    \includegraphics[height=3.5cm]{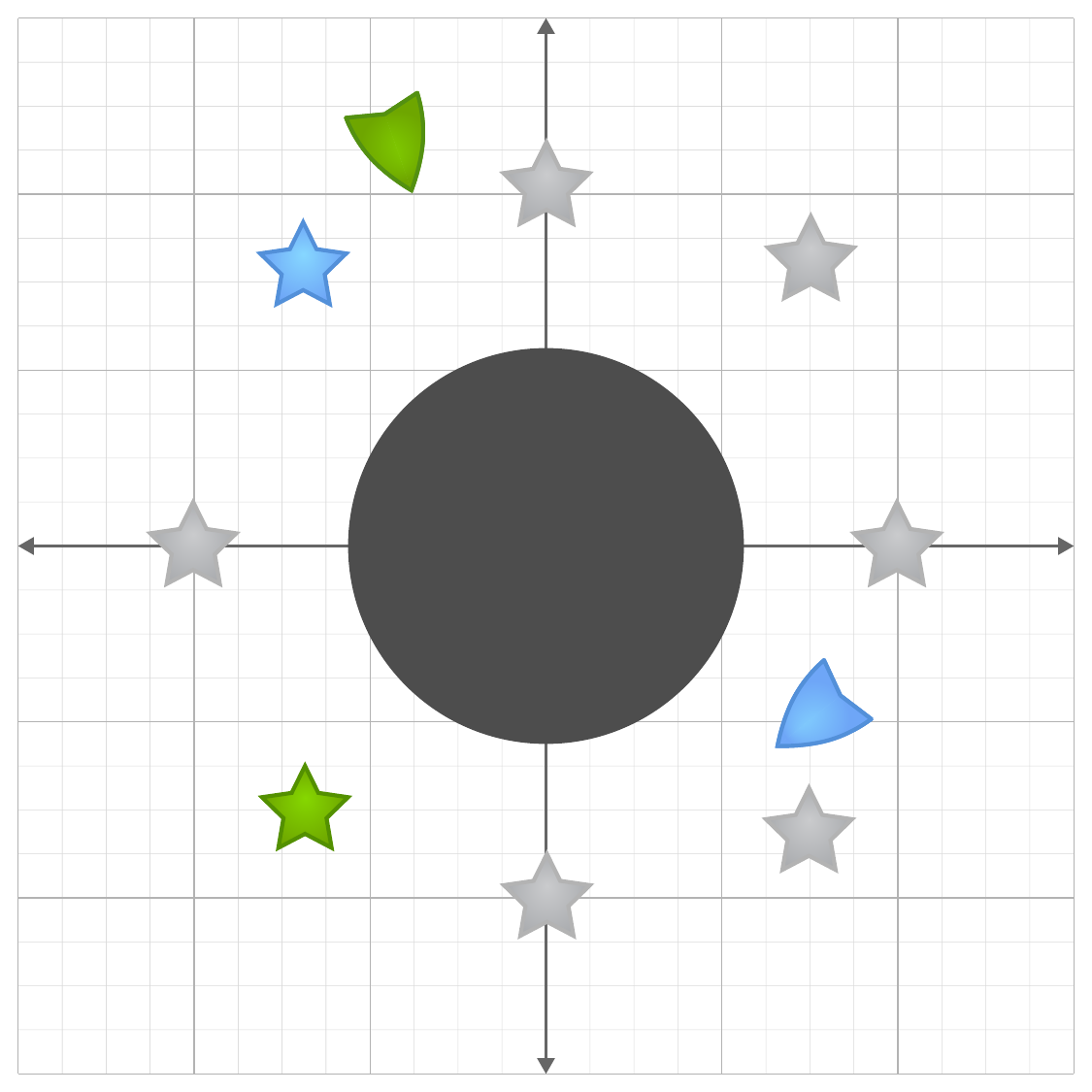}
    \caption{Illustration of the\\driving environment (Section \ref{sec:envs}).}
    \label{fig:pit_env_illustration}
\end{subfigure}%
\caption{{(a)} Agents architecture diagram for two agents (top and bottom) with communication. {(b)} Illustration of the gridworld toy environment with two agents, depicted with blue and green circles on a 5x5 grid. The green and blue stars indicate the locations of each agents' respective goals. The cloud thought bubble depicts the world as the green agent observes it. Note, each agent does not see their own goal, and they are off-by-one square in their knowledge of each other's goal. {(c)} Illustration of the driving environment with two agents. The dark circle in the centre is the `pit'; the region in which large negative penalties are given when agents enter. The stars indicate goal locations, which can spawn in one of eight locations (the unused locations indicated by the greyed-out stars). The continuous state space is indicated with the grid axes. The agents are represented by arrowheads indicating their current position and direction. Again, here agents do not observe their own goals and therefore need to communicate, but unlike the gridworld, they do have perfect knowledge of the other agent's goals.}
\label{fig:arch_and_env}
\end{figure*}

Imitation Learning (IL) is a form of machine learning in which expert demonstration data is used to construct a policy for solving a task. The data is typically in the form of an observation $o$ and action $a$, and the imitation learning problem is posed as supervised classification learning, e.g. optimising $\theta$ to minimise the difference between $a$ and $\hat{a} = \pi_\theta(o)$. See \citet{hussein_imitation_2017} for a more comprehensive review of these methods.


\section{Cooperative Language Acquisition}

In this section, we introduce our definition of a \textit{Cooperative Language Acquisition Problem} (CLAP). A CLAP can be formulated from the simplest case involving a preexisting \textit{target community} of two agents, denoted $\Pi = \{A, B\}$, that achieve some non-trivial performance in a Dec-POMDP-Com $\mathcal{M}$.
Consider an interaction at time $t$ in which $A$ is speaking and $B$ is listening.
While making the observation $o_t^s$, the speaker emits a message $m_t$.
Then, while observing $o_t^r$ and $m_t$, the receiver takes the action $a_i$ in the environment.
We are given access to $\mathcal{M}$ and a dataset of such interactions, denoted $(o_t^s, m_t, o_t^r, a_t^r) \in \mathcal{D}$, and our task is to construct a \textit{joiner agent} $\joiner$.
In this paper, we focus on a specific case of the task in which we aim to replace a target agent in $\Pi$ (so there is always a fixed number of players), which we call CLAP-Replace.
The agent $\joiner$ should be able to take on the role of a target agent (e.g. either $A$ or $B$) and successfully communicate with the other, while also acting in the environment to maximise the cooperative reward.
The joiner will be evaluated zero-shot, i.e. \textit{on the first joining event}, so all learning must be done beforehand.

The emphasis of the dataset is on learning the communication protocol as this is assumed to be the key aspect of the community's strategy that cannot be learned independently from observing their behaviour.
As the communication symbols have no prior semantics, any protocol could be equally successful if the same meanings were assigned to a different permutation of the symbols.
On the other hand, there may be many `environment-level' behaviours that can be learned without needing to know the specific strategies of a given community.
We will discuss disentangling these factors in Section \ref{sec:env_level_vs_comm}.

\subsection{Problem Definitions}

The cooperative language acquisition task is to construct a \textit{joining agent} $\joiner$ that learns from observing a target community $\Pi$. When $\joiner$ replaces an agent in $\Pi$, we have a CLAP-Replace task:

\begin{problem}[CLAP-Replace]
    Suppose there exists a Dec-POMDP-Com $\mathcal{M}$ and a set of policies $\Pi = \{\pi_i\}$ trained in $\mathcal{M}$.
    Given a target policy $\pi_k \in \Pi$ to replace, a dataset $\mathcal{D}_{k}$ of interactions $(o_t^s, m_t, o_t^r, a_t^r)$ in which $\pi_k$ is either a speaker or listener, the task is to construct a policy $\pi_\joiner$ such that:
    \begin{align}
        \pi_{\joiner} \in \argmax_{\pi'} R(\{\pi'\} \cup \Pi^{-k})
    \end{align}
    Where $\Pi^{-k} = \Pi \setminus \{\pi_k\}$. In other words, maximise the team rewards when the joiner replaces $k$.
\end{problem}

This can be decomposed into three related problems; a \textit{forward communication} (signalling), a \textit{backward communication} (listening), and \textit{acting}. More precisely, $\joiner$ needs to learn to send messages that maximise the expected sum of rewards for listening agents and interpret messages to maximise the rewards of its trajectory. But the agent also needs to learn to utilise information, both communicated and observed, for selecting actions.

If we were allowed to interact with the target community $\Pi$ before evaluation, we could apply standard reinforcement learning tools. However, we are interested in settings in which disrupting the community and jeopardising performance to learn is not acceptable, so $\joiner$ needs to perform well on the first actual joining episode.

\subsection{Disentangling Environment-Level and Communicative Competencies}
\label{sec:env_level_vs_comm}

We assume that teams of agents can achieve a certain level of performance in the environment by two categories of competencies: (1) `environment-level' skills, and (2) communication strategies that rely on established conventions.
The key distinction we aim for is that the former should be learnable independently of observing a target community's behaviour.
We start by defining that the team achieves an average total reward of $R$ when acting together in the environment and communicating.

Measuring the \emph{communicative competency} (1) is relatively easy: if communication is blocked and the team achieves a lower average total reward $R' < R$, and by assuming that the agents are engaged in positive listening and signalling, we can attribute this drop in performance to a lack of communication.

\begin{figure}
    \centering
    \includegraphics[width=\linewidth]{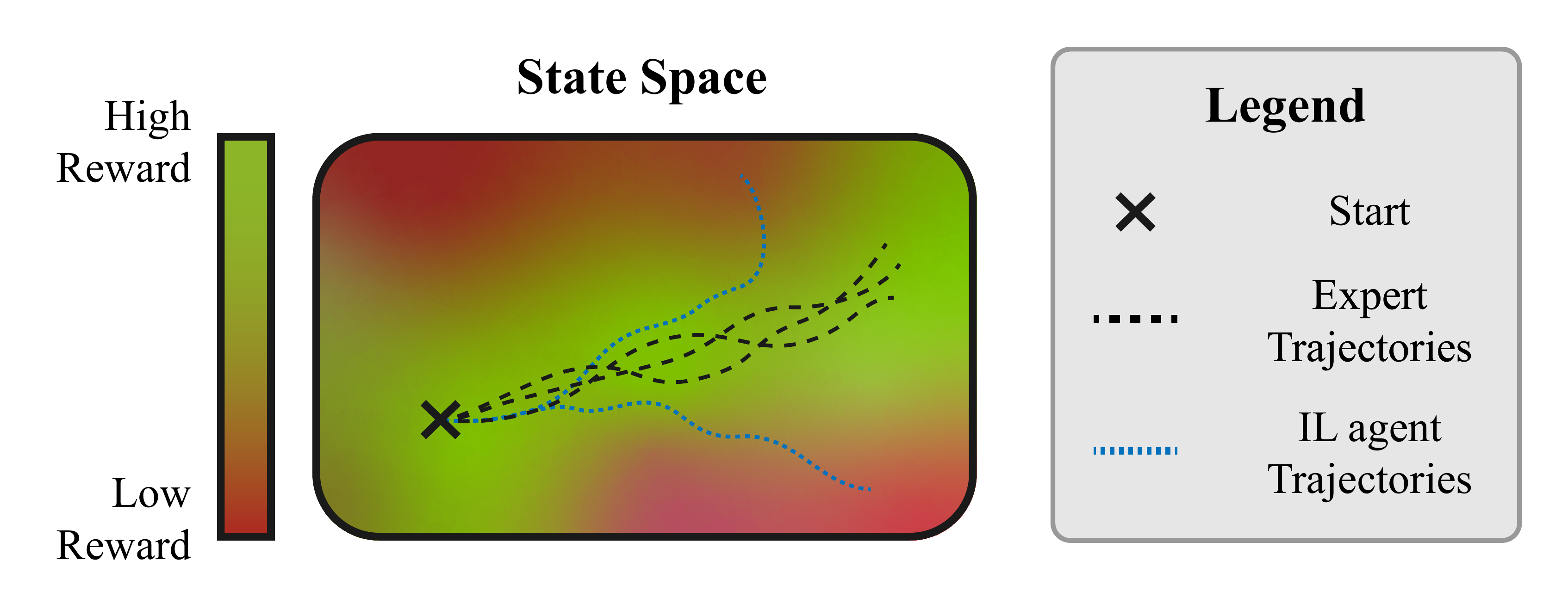}
    \caption{Illustration of the problem of compounding errors for Imitation Learning. When IL agents exit the expert state distribution they are unable to recover.}\label{fig:il-compounding-errors}
    \label{fig:enter-label}
\end{figure}

We evaluate the influence of \emph{environment-level competencies} (2) by similarly hampering them, and investigating whether this leads to a drop in performance.
We will assume that each agent's observation can be partitioned into `private' (or `local') and shared (or `global') components: $o^i_t = (l^i_t, g_t)$, where $g_t$ is observed by both speakers and listeners, but $l^i_t$ is private to $i$. 
To assess the contribution of environment-level skills we restrict the environment-level policy by removing private information, $o'_t = (\mathbf{0}, g_t)$, and replacing each $\pi_i$ with $\pi'_i(\mathbf{o}_t, M_t) = (\pi^e_i(o'_t, M_t), \pi^c_i(o_t))$.
We can then attribute a change $R'' < R$ to the agent being prevented from exercising environment-level skills. 
To understand this intervention, consider the following points:

\begin{figure*}
    \centering
    \includegraphics[width=\linewidth]{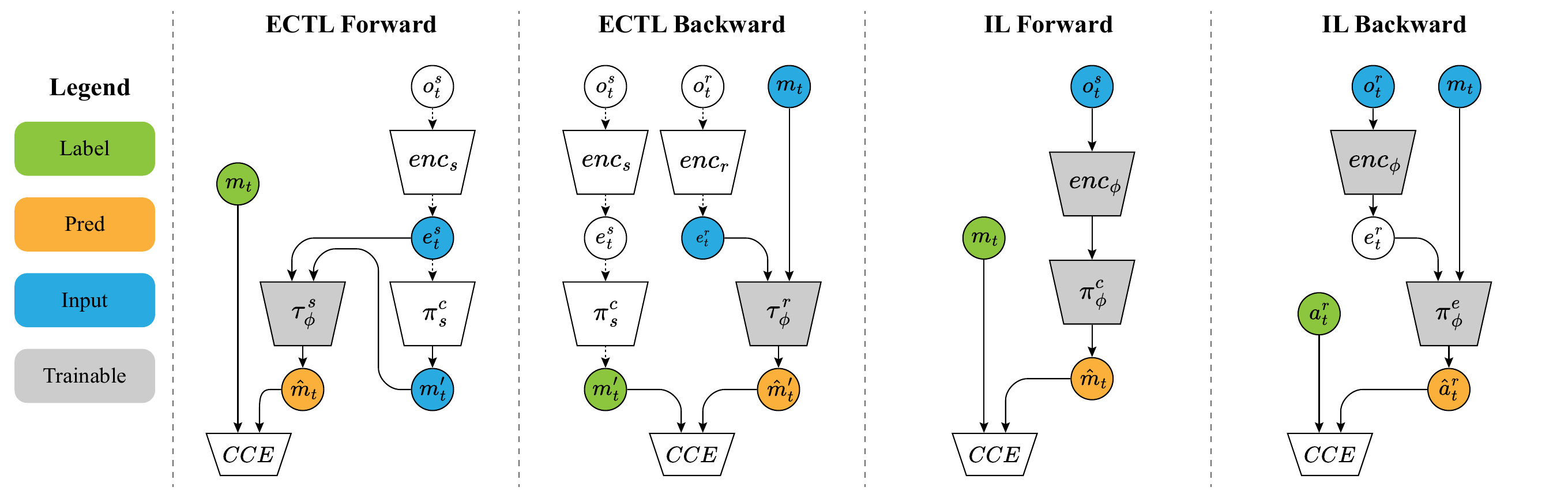}
    \caption{Training architecture diagrams for the \textit{Imitation Learning} (IL) and \textit{Emergent Communication pretraining and Translation Learning} (ECTL) methods. The forward and backward problems for each method are solved with supervised learning. Circles are variables and trapezoids are functions. Dotted lines indicate that gradients are blocked from backpropagating along a path. For the ECTL diagrams the $enc$ and $\pi^c$ functions are learned during the emergent communication pretraining phase. The variables $o^s_t, o^r_t, m_t, a_t$ are the sender/receiver agent observations, the sender's message, and the receiver's action from the dataset of interactions collected from the target community.}
    \label{fig:training-arch-diagrams}
\end{figure*}

\begin{enumerate}
    \item
    We hamper environment-level skills by blocking private information from $\pi^e$, but an agent must still be allowed to receive messages as this may provide vital information for succeeding based on communication skills. Thus, $\pi^c$ for the speaker must not be intervened with.
    \item Communication may rely on shared information, so removing the context by blocking this information may interfere with measuring environment-level skills.
    \item Information private to a recipient cannot have been used by the speaker to create the messages, so if a drop in performance is seen when it is removed from $\pi^e$, this can only be due to a loss in the recipient's capacity to use that private information to contribute to the team's performance through its actions (as opposed to its contributions through sending messages).
    Recall, we are only obfuscating this information from $\pi^e$; $\pi^c$ still sees it.
    \item If no drop from blocking private information to $\pi^e$ is observed, it implies one of two cases: (a) all of an agent's contribution to the collective performance is the result of following direct orders from a speaker. Or, (b) an agent's contribution is entirely contingent on global information.
\end{enumerate}

Note that following 4b, the absence of a drop does not imply that the team does not possess environment-level competencies, but the presence of a drop does.

\section{Methods for Constructing Joiners}

In this section, we introduce two methods for constructing joiners for the CLAP-Replace task.
The first naively applies imitation learning, and the second pretrains agents using emergent communication and then translates the learned communication protocol to the target community's protocol.
Each method is decomposed into forward and backward problems, posed as supervised learning tasks.

\subsection{Imitation Learning (IL)} \label{sec:il-clap}

\begin{figure*}[t]
\centering
\captionsetup[subfigure]{justification=centering}
\begin{subfigure}{.48\textwidth}
    \centering
    \includegraphics[height=3cm]{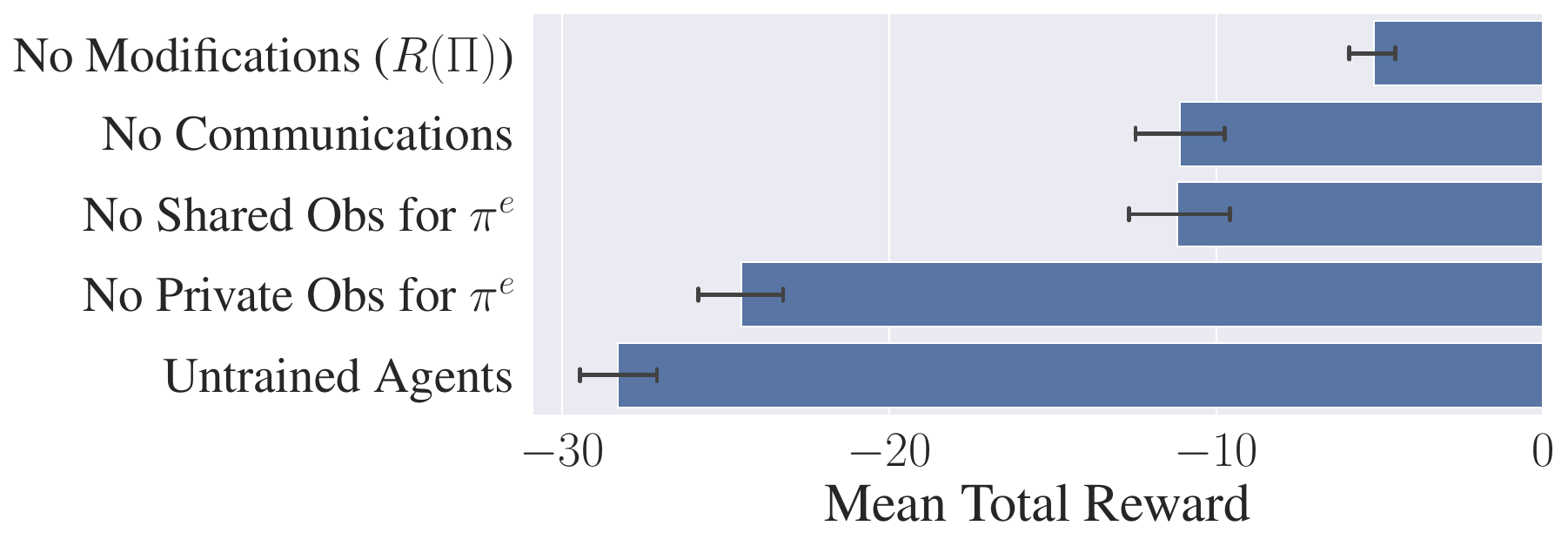}
    \caption{Ablated agents in the target community.}
    \label{fig:ablation_evaluation_results}
\end{subfigure}%
\hfill
\begin{subfigure}{.48\textwidth}
    \centering
    \includegraphics[height=3cm]{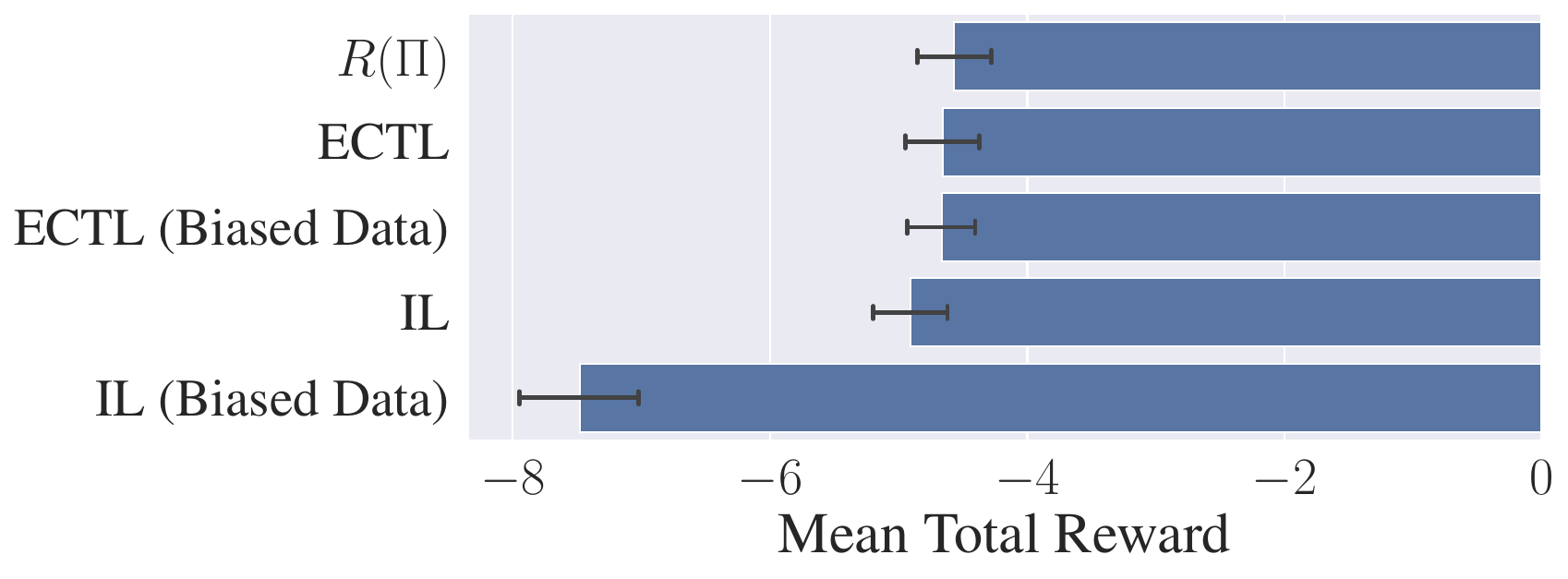}
    \caption{Mean reward for different teams.}
    \label{fig:joining-evaluations}
\end{subfigure}%
\caption{Performance results from cases in which (a) the team is formed of a variety of agents, comparing ECTL to imitation learning (IL) with unbiased and biased data from the gridworld environment. And (b) the target community agents ablated in different ways. All results are from 500 evaluation episodes and the error bars show means within 95\% confidence intervals.}
\label{fig:arch_and_env}
\end{figure*}

In the set-up for CLAP-Replace we are given a dataset $(o_t^s, m_t, o_t^r, a_t^r) \in \mathcal{D}_k$ of speaker and listener observations, messages, and actions taken by the receiver.
The simplest baseline solution to this problem is to apply imitation learning separately to the signalling and listening problems.
The dataset is partitioned into two datasets: a signalling dataset $(o_t^s, m_t) \in \mathcal{D}^s_k$ where $k$ is the speaker and a listening dataset $((o_t^r, m_t), a_t^r) \in \mathcal{D}^r_k$ where $k$ is the receiver.
These datasets are structured as input-label tuples to learn communicative and environmental-level policy factors ($\pi_{il}^c$ and $\pi_{il}^e$) of the overall imitation policy $\pi_{il}$.
These are learned with a categorical cross-entropy ($CCE$) loss between predicted and actual labels. We will denote this \textit{imitator} joiner $\joiner_{il}$. In Figure~\ref{fig:training-arch-diagrams}, $\pi^c_{il}$ and $\pi^e_{il}$ are composed of encoder and communication/action heads, with the same architecture as in Figure~\ref{fig:arch_diagram}.

\subsection{Emergent Communcation Pretraining and Translation Learning (ECTL)} \label{sec:ectl}

\begin{figure*}[t]
\centering
\begin{subfigure}{.45\textwidth}
    \centering
    \includegraphics[height=5cm]{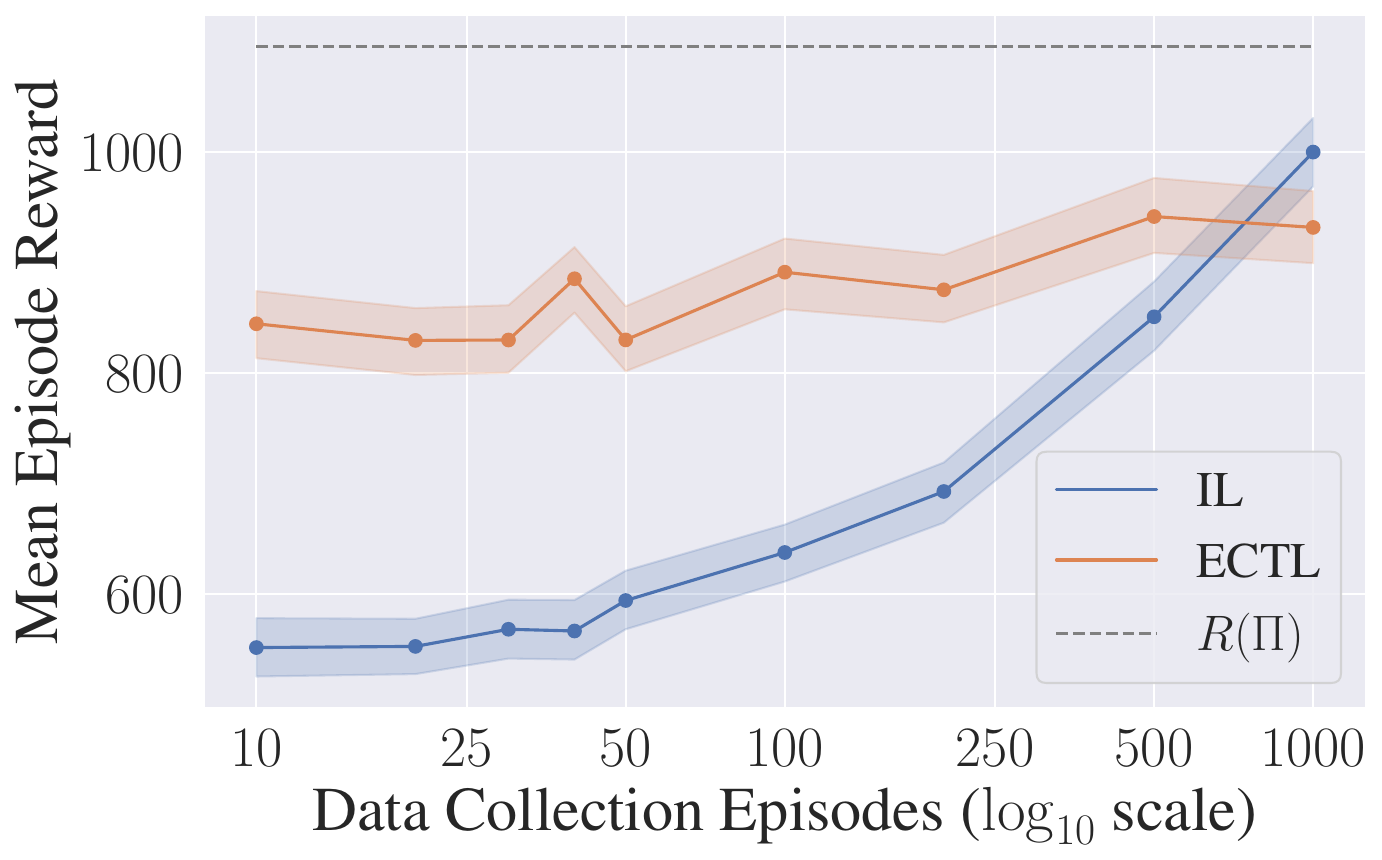}
    \caption{
        The mean CLAP-Replace performance on the Driving Game (No Pit) for ECTL and IL,
        against the number data collection episodes ($N_{collect}$) from the target community.
    }
    \label{fig:reward_vs_collect_eps}
\end{subfigure}%
\hspace{0.025\textwidth}
\begin{subfigure}{.45\textwidth}
    \centering
    \includegraphics[height=5cm]{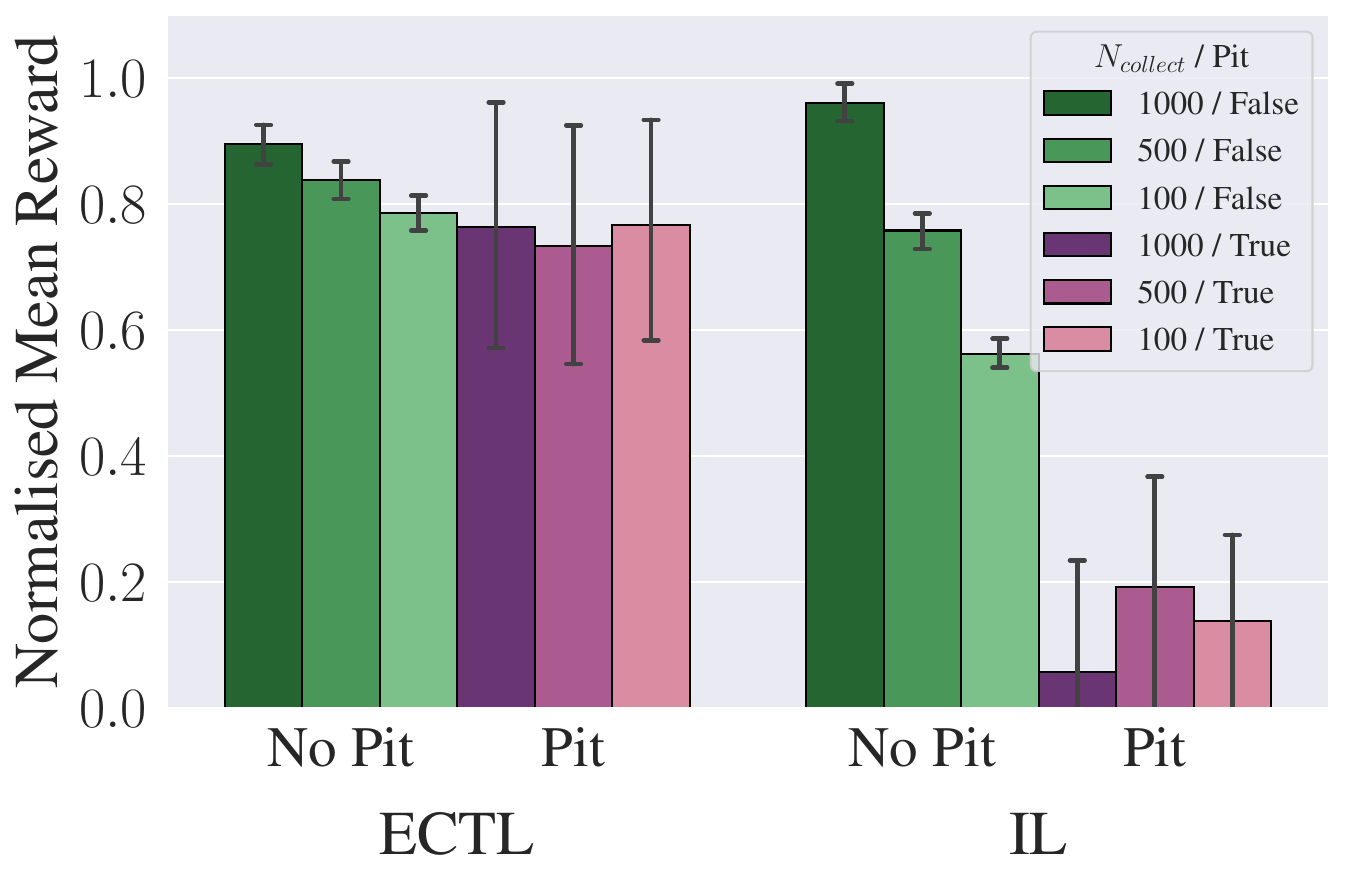}
    \caption{Mean reward for each method normalised by the original team mean rewards. Shows the impact of the pit on CLAP performance, highlighting IL's relative fragility.}
    \label{fig:jtc-pit-vs-not-pit}
\end{subfigure}%
\caption{Comparisons between IL and ECTL on the Driving Communication environment.}
\label{fig:arch_and_env}
\end{figure*}

A well-known issue with imitation learning agents is that they can be brittle given the natural biases in the expert demonstration data \citep{kumar_should_2022}.
This problem is illustrated in Figure~\ref{fig:il-compounding-errors}.
Especially in more complex domains, experts typically stay in the regions of state space in which they get high rewards, which may only be a small portion of the possibilities.
If an imitator makes a mistake when attempting the task itself, it may enter into unseen territory.
Thus errors can compound as the agent has not learned what to do and makes more mistakes, leading to degraded team performance in a multi-agent cooperative setting.

This leads to the idea that the joiner could explore the state space of $\mathcal{M}$ before joining $\Pi$, and thereby become a more reliable cooperator.
In general, pretraining may allow the agent to learn environment-level skills that could be transferable to cooperating with any new team.
To this end, we propose the method \textit{Emergent Communication pretraining and Translation Learning} (ECTL).
The first step of ECTL is to (pre)train a set of agents $\Pi'$ from scratch in $\mathcal{M}$ that we refer to as the \textit{EC training community}, where for each agent in $\Pi$ there is a corresponding agent in $\Pi'$.

The agents in $\Pi'$ are composed of three components, illustrated in Figure \ref{fig:arch_diagram}: an \textit{observation encoder} $enc_\theta$, a \textit{communications head} $\pi^c_\theta$, and an \textit{action head} $\pi^e_\theta$, parameterised by $\theta$, and shared amongst the policies $\Pi'$.
For brevity and to be consistent with the notation in previous sections we will omit the encoder from our notation, and write $\pi^c_\theta(o)$ instead of $\pi^c_\theta(enc_\theta(o))$.
This training process could use any viable multi-agent reinforcement learning algorithm, and which is most suitable depends on the properties of the underlying environment and the number of agents.
In Section \ref{sec:train_target_comm}, we will outline the specific methods we used for our experiments.

Agents in the EC training community learn to cooperate via an emergent communication protocol over their message alphabet $\Sigma'$.
The next step in ECTL towards building the joiner agent for $\Pi$ is to \textit{translate} this protocol to the one used by the target community.
For a CLAP-Replace task with target agent $\pi_k\in\Pi$, we select the equivalent pretrained $\pi'_k \in \Pi'$ to use a starting point for \textit{translation learning}.
We will refer to $\pi'_k$ as the \textit{EC pretrained agent}.

Once again, the target community data $\mathcal{D}_k$ is transformed into a signalling dataset $(o_t^s, m_t) \in \mathcal{D}_k^s$ and a listening dataset $((o_t^r, m_t), m'_t) \in \mathcal{D}^r_k$.
Note that the ECTL signalling dataset is identical to the signalling dataset used for imitation learning, but the listening dataset is different.
Instead of the label being an action, it is a message $m'_t$ from the EC pretraining message space $\Sigma'$.
To construct these labels we use the speaker's observation from $\mathcal{D}_k$ and compute the message that EC pretrained agent would have emitted that same situation, i.e. $m'_t = \pi^c_\theta(o^s_t)$.

From these data, we learn \textit{translation functions} for signalling and for listening. A separate translation function can be learned for each communication channel if there are multiple possible senders and/or receivers for the target agent, or two models (one for each CLAP sub-problem) can be used by providing the sender/receiver identifiers as input. For our experiments, we will use the latter approach. However, for simplicity in the following descriptions of how these functions are trained, we will assume one sender and receiver for the target agent and omit agent identifiers. Additionally, the training architectures for the translation functions are illustrated by the two leftmost diagrams of Figure~\ref{fig:training-arch-diagrams}.

A signalling translation function $\tau^s_\phi:\Omega_k\times\Sigma'\rightarrow\Sigma$ maps from the EC training community's communication protocol to the target community's protocol, and is parameterised by $\phi$.
Given an observation $o_t^s \in \Omega_k$ and message $m_t\in \Sigma$ from $\mathcal{D}^s_k$, we can compute the message $m'_t = \Sigma'$ that the agent $\pi'_k \in \Pi'$ would send in that situation. The translation function is trained to predict the demonstrator agent's message:
\begin{align}
    &\phi^* \in \argmin_\phi \sum_{(o_t^s, m_t) \in \mathcal{D}^s_k} CCE(m_t, \hat{m}_t) \\
    &\text{where}~\hat{m}_t = \tau^s_\phi(o^s_t, m'_t), \quad m'_t = \pi^c_\theta(o_t^s)
\end{align}

A listening translation function $\tau^r_\psi:\Omega_k\times\Sigma\rightarrow\Sigma'$ maps from the target community's communication protocol to the EC training community's protocol, and is parameterised by $\psi$.
Note that, in general, $\tau^r$ and $\tau^s$ need not be inverses of one another, as two agents can use arbitrarily different protocols in each direction.

To train $\tau^r_\psi$, we use the message that was received by the listener (the target agent) $m_t$, the listener's observation $o^r_t$, and the speaker's observation $o^r_t$.
The translation function takes the observation and the message and produces a new message in the EC training community's message space: $\tau^r_\psi(o^r_t, m_t) = \hat{m}'_t \in \Sigma'$.
To get the ground-truth label for this prediction, we find the message that the EC pretrained agent would have sent in the target community speaker's situation.
The result is the following optimisation criterion:
\begin{align}
    &\psi^* \in \argmin_\psi \sum_{((o_t^r, m_t), m'_t) \in \mathcal{D}^r_k} CCE(m'_t, \hat{m}'_t) \\
    &\text{where}~\hat{m}'_t = \tau^r_\psi(o^r_t, m_t), \quad m'_t = \pi^c_\theta(o^s_t)
\end{align}

Finally, we can put together these pieces into a final \textit{translator} joiner agent $\joiner_{ectl}$ composed of the following communication and environment-level policy factors:
\begin{align}
    \pi^c_{ectl} =  \tau^s_\phi \circ \pi^c_\theta \quad\text{and}\quad \pi^e_{ectl}(o, m) = \pi^e_\theta(o, \tau^r_\psi(m))
\end{align}
So when the joiner agent receives a message, it uses the listening translation function to predict the message that it would have received from the equivalent agent in its EC training community.
When speaking, the agent first computes the message that it would have sent in its original message space, and then passes that through the signalling translation function before finally sending it.

\section{Experiments}
\subsection{Environments} \label{sec:envs}

In order to empirically investigate IL and ECTL for solving CLAP-Replace tasks, we created two environments. Firstly, a simple gridworld toy environment in which $N$ agents cooperate by communicating goal information. Secondly, a `driving' communication problem in which agents must navigate a continuous space to reach a goal, while potentially avoiding a pit in their way. Again, the goal location is known by another agent so they need to communicate to solve the problem.

\textbf{Goal Communications Gridworld.}
Each agent has a goal square in the grid that they need to reach that changes each episode.
No agent observes its own goal unless it is within one tile of it, but at all times it observes a `close guess' (a location within one tile) of the goal of another of the agents in the game.
The environment is illustrated in Figure \ref{fig:toy_env_illustration}.

\textbf{Goal Communications Driving Game.} As discussed in Section \ref{sec:ectl}, IL agents can be brittle when regions of the state space are not present in the demonstration data.
To investigate this problem, we introduced this `driving' environment in which the agent steers and accelerates a body in a continuous grid (Figure \ref{fig:pit_env_illustration}).
This environment has two settings defined by the presence or absence of a circular region in the centre of the world where a large penalty is applied for every agent within the region.
The agents must navigate to one of eight fixed goal locations, known by their partner and selected at random for each episode.

\subsection{Creating Target Communities}
\label{sec:train_target_comm}

Target communities $\Pi$ of three agents in the gridworld environment and two agents in the driving environment were trained using \textit{Multi-Agent Proximal Policy Optimisation} (MAPPO) \citep{yu_surprising_2022}.
For the policy networks we used the architecture in Figure \ref{fig:arch_diagram} with parameter sharing between agents. 
The value network used a concatenation of all of the agents' observations as input, meaning that we rely on centralisation for training.
The value network did not share any parameters with the policy networks.
A Gumbel-Softmax \citep{jang_categorical_2017, maddison_concrete_2017} communication channel function was used on communication channels.
Further training details can be found in Appendix \ref{app:training_params}.

\subsection{Ablating Target Community Agents}

\begin{figure*}[t]
\centering
\begin{subfigure}{.25\textwidth}
    \centering
    \includegraphics[height=3.5cm]{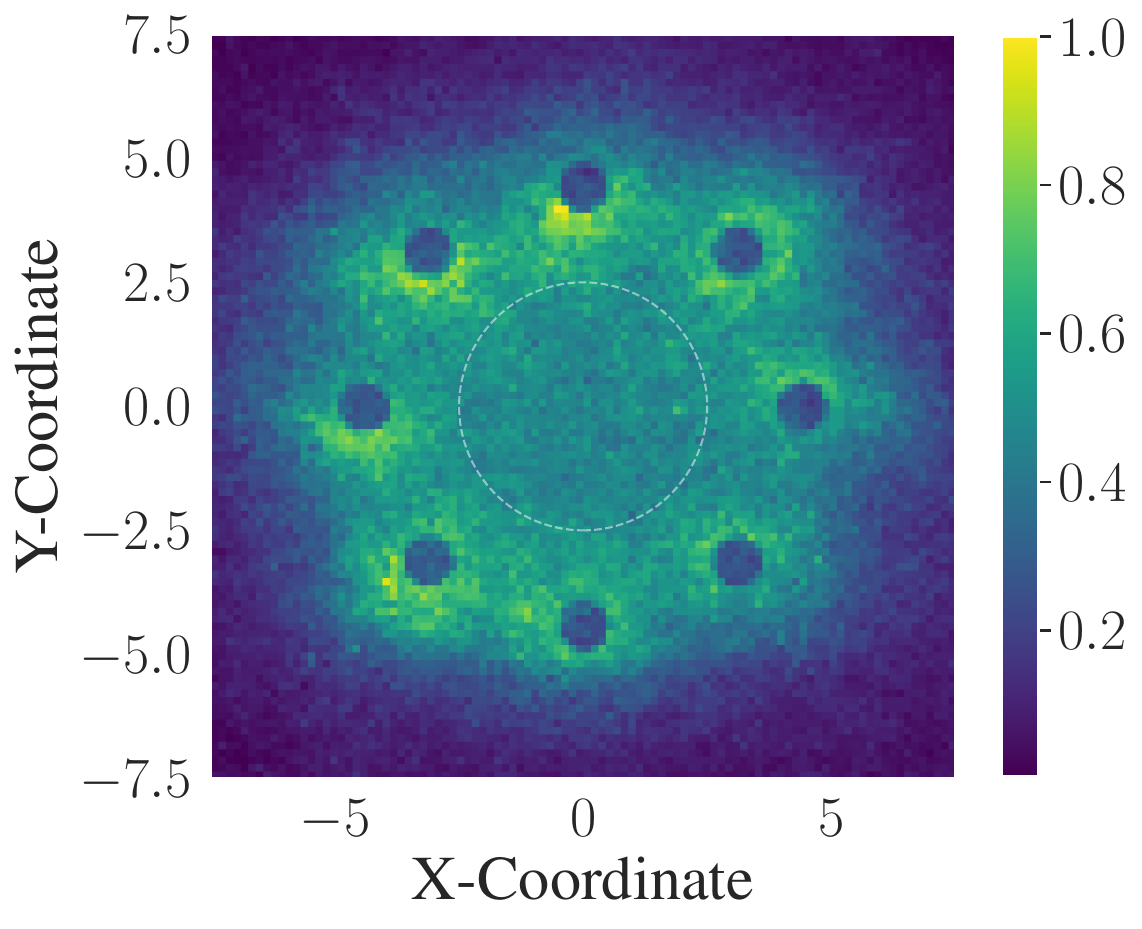}
    \caption{}
    \label{fig:target_community_driving_data_without_pit}
\end{subfigure}%
\begin{subfigure}{.25\textwidth}
    \centering
    \includegraphics[height=3.5cm]{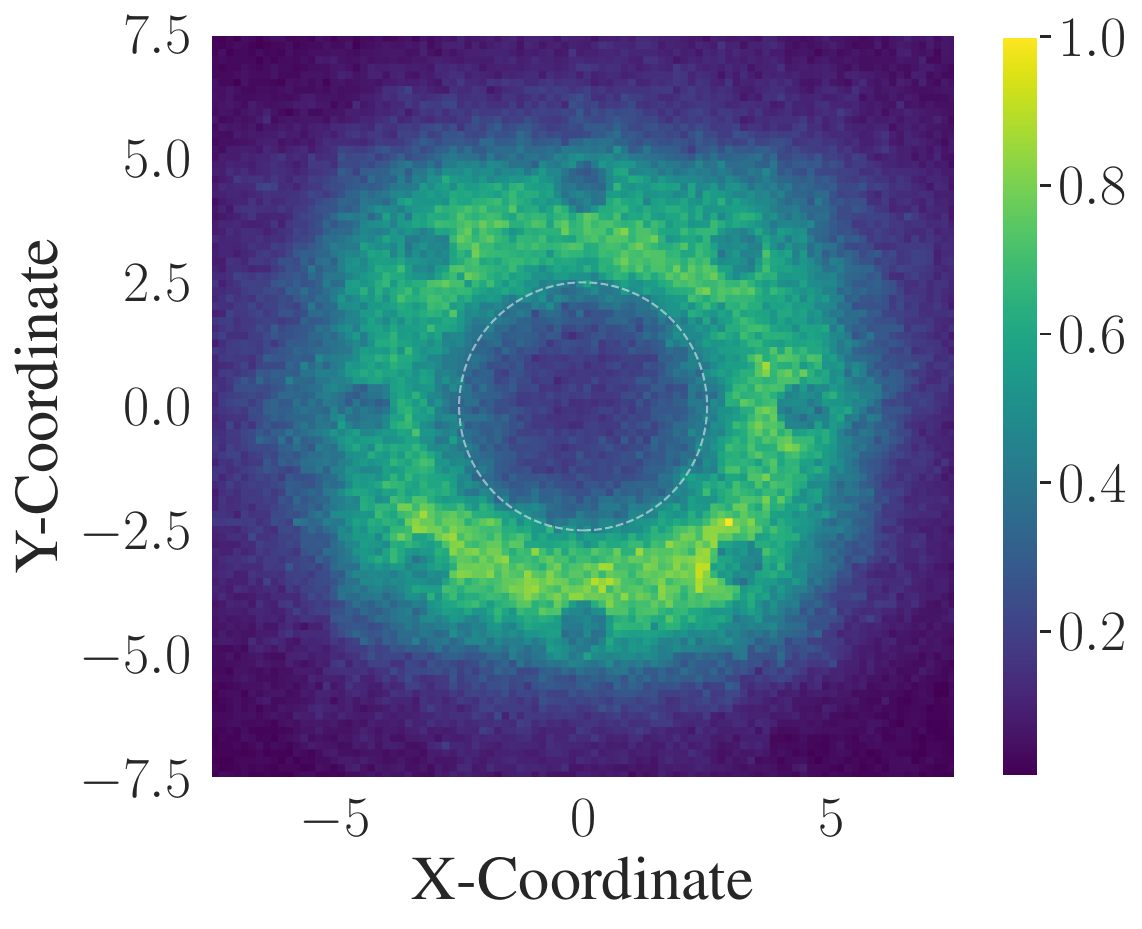}
    \caption{}
    \label{fig:ectl_driving_data_with_pit}
\end{subfigure}%
\begin{subfigure}{.25\textwidth}
    \centering
    \includegraphics[height=3.4cm]{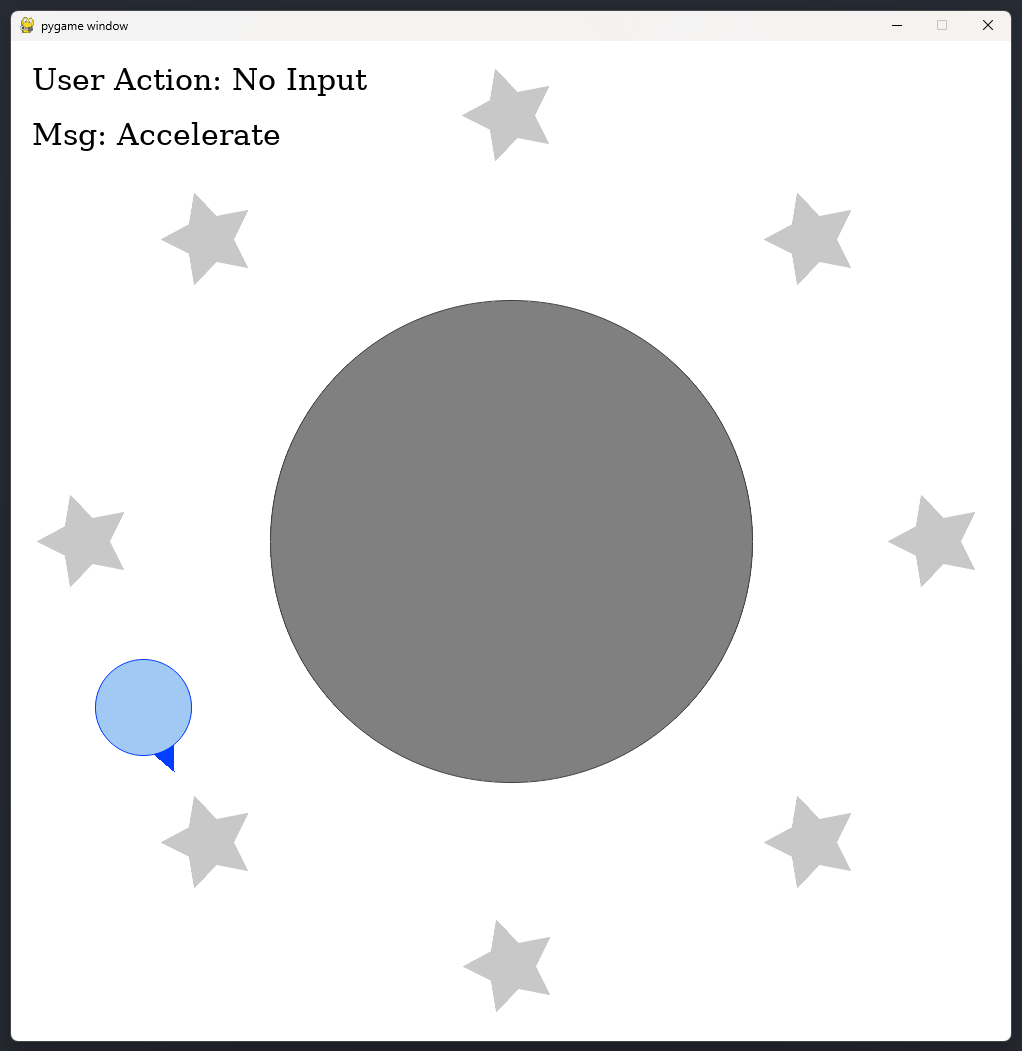}
    \caption{}
    \label{fig:interactive-driving-ui}
\end{subfigure}%
\begin{subfigure}{.25\textwidth}
    \centering
    \includegraphics[height=3.2cm]{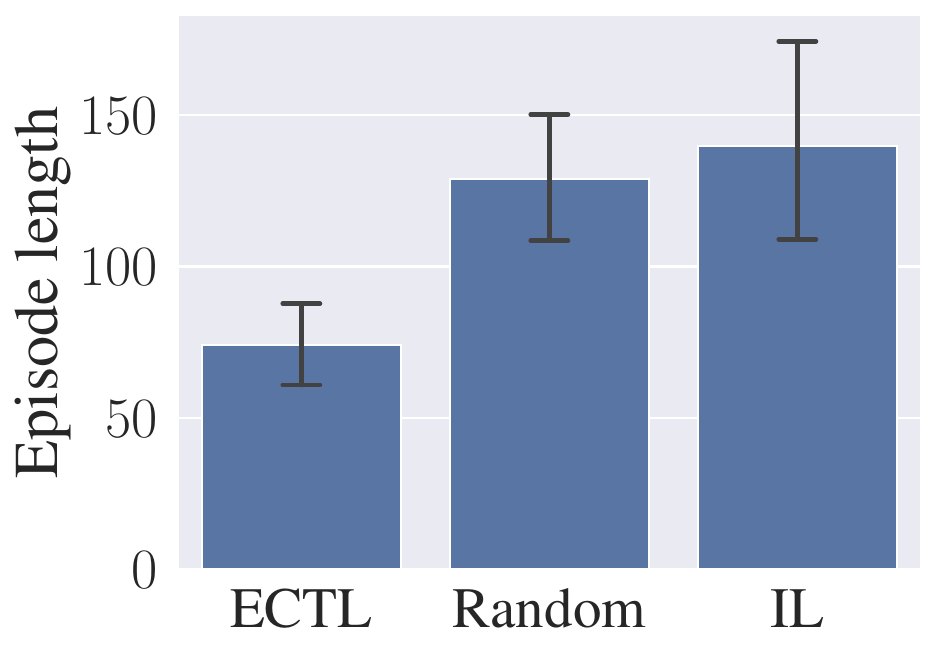}
    \caption{}
    \label{fig:playing_with_clap_agents_episode_lens}
\end{subfigure}%
\caption{(a,b) Two-dimensional histograms visualising how much time the agents spend in the different regions of the environment before reaching their goals. Left and right show agents trained with/without pit penalties. The radius of the pit is indicated with the dashed circle outline. We can see that when the pit is present agents avoid that region. The goal locations are also visible as we are not plotting positions after the goals are reached. (c) The UI for the human to play the game as a listener. The top-left of the screen shows messages from the other player. (d) The mean episode lengths when we played this game with different agents. The goal is for us to navigate to our goal location as quickly as possible (one of the grey stars) using the messages. We see that only the ECTL agent was able to effectively communicate.}
\label{fig:arch_and_env}
\end{figure*}

The gridworld environment was designed such that the agents need to use a combination of communicative and individual skills in order to succeed.
To demonstrate how these skills contribute to the performance, we evaluate ablated versions of the agents with the interventions discussed in Section \ref{sec:env_level_vs_comm}.
Figure \ref{fig:ablation_evaluation_results} shows the results of these experiments.
At the top, we show the original unmodified performance for reference ($R(\Pi)$).
The next two bars, appearing roughly equal in size and variance, correspond to blocking communications (bar 2) and `global'/`shared' information from $\pi^e$ (bar 3).

\subsection{Gridworld CLAP-Replace}

We train ECTL and IL agents after collecting interaction datasets from $N_{collect}=100$ episodes of $\Pi$ acting in the environment.
See Appendix \ref{app:training_params} for details on training hyperparameters and learning curves.
Figure \ref{fig:joining-evaluations} shows the results from forming teams of different compositions in the gridworld environment.
We see that ECTL and IL solve the CLAP-Replace task by achieving the equivalent team performance as the original team ($R(\Pi)$). 

Both methods have roughly the same performance, which is due to the fact that there are not opportunities for errors to compound.
The imitator will have learned the experts' behaviour from any state as the agent starting positions and goal positions are sampled from a uniform distribution over all grid tiles.
Therefore, if an imitator makes a mistake from any given position, it will have seen during training how to recover in the next position.

So to investigate the effects of unseen states we bias the training data by only including cases where the target agent was in one of the first two columns of the grid.
The results are also shown in Figure~\ref{fig:joining-evaluations}, and we see that while ECTL maintains its performance, IL significantly degrades.



\subsection{Driving Game CLAP-Replace}

Moving on from the gridworld environment, we then performed a series of experiments in the driving game.
First, we trained four teams on the environment, with two agents per training team. Two of the teams with the pit, and the other two without the pit.
Figures \ref{fig:target_community_driving_data_without_pit} and \ref{fig:ectl_driving_data_with_pit} visualise where in the environment these agents spend their time (before reaching their goals), illustrating how the expert demonstrations may not cover the pit region of state space.

In Figure \ref{fig:reward_vs_collect_eps} we see the effect of the number of data collection episodes on the driving game CLAP-Replace performance without the pit.
We find that the ECTL performance starts notably better than IL, and can do well with very little data.
However, as more data is gathered, eventually IL scales better than ECTL.
Yet, Figure \ref{fig:jtc-pit-vs-not-pit} complicates this picture.
Here we see the normalised reward comparisons between ECTL and IL showing the effect of the pit on performance.
The pit does not have much impact on ECTL, but the performance of IL drops by a large margin, even for the case with 1000 data collection episodes -- the setting in which IL outperforms ECTL when the pit is not present.

\subsection{Translating to an Interpretable Communication Protocol}

After demonstrating that IL and ECTL could successfully solve CLAPs for target communities of artificial agents, we investigated whether or not the same methods could apply to data generated by humans.
We developed an interactive UI through which a user could simultaneously control two agents in the driving game while observing the entire game state (i.e.\ both agent's goals).
The user observes the world through a top-down view like in Figure~\ref{fig:pit_env_illustration}, with both agents, the goal locations, and the pit visible.
To control the first agent in the game, the user presses the `W', `A', and `D' keys to accelerate, turn anti-clockwise, and turn clockwise respectively.
Likewise, the user controls the second agent with the `Up Arrow', `Left Arrow', and `Right Arrow' keys.
Additionally, to make it easier to control both agents at the same time the frame rate of the game was reduced to 2 frames per second.
This was then used to collect data from 70 episodes.
Each row of data comprised the actions that the user took for each agent and the agent-specific observations.
Formally, each row of data had the form $(o^1_t, o^2_t, a^1_t, a^2_t)$, where $t$ is the time step that the data was collected, $o^i_t$ and $a^i_t$ are the observations and actions for each agent $i$.

In its raw form, this data cannot be used to train CLAP agents as it lacks messages.
To remedy this we augmented the data by constructing messages on the principle that each agent was following direct instructions from the other, so each agent $i$'s action $a^i_t$ became agent $j$'s message $m_t=a^i_t$ to agent $i$.
Now each row of the data can be transformed into two CLAP-Replace datasets rows, one dataset for each agent that could be replaced: $(o^1_t, m_t=a^2_t, o^2_t, a^2_t)$ and $(o^2_t, m_t=a^1_t, o^1_t, a^1_t)$.

Finally, we removed all rows where the user was not inputting one of the three movement actions (turn clockwise/anti-clockwise or accelerate) as it was hard to coherently press keys for both agents with exact simultaneity. 
Furthermore, as each agent had the same observation space we could take each collected episode and construct data for an alternative episode where the agent roles were reversed, resulting in a final dataset of 140 episodes.
Formally, for each row of the original raw collected data $(o^1_t, o^2_t, a^1_t, a^2_t)$, we construct another row $(o^2_t, o^1_t, a^2_t, a^1_t)$.

With this data, ECTL and IL agents could now be trained in the same manner as done for the MAPPO target communities.
However, due to the lack of a well-defined target community, the agents could not be evaluated in the same way.
So instead, we focused on specifically evaluating the signalling capabilities of these agents with another UI that allows a user to view their agent, the possible goal locations, and a message from one of the artificial agents (Figure~\ref{fig:interactive-driving-ui}).
As the message space is the action space for the user, it is inherently interpretable and could be rendered to the user as one of the strings `Accelerate', `Turn Anti-Clockwise', or `Turn Clockwise'.
We then set up the UI to randomly switch the messenger agent each episode from a pool of agents comprised of an ECTL agent, an IL agent, and an agent that sends a random message.
Thus the user was always unaware of which agent they were playing with.
This UI runs at 10 frames per second, but the message only updates every half second (every 5 frames).
This is to prevent it being too obvious which is the random agent as the messages would change much more frequently than for the other agents.
Additionally, it gives the user more time to read and react to the messages.

Figure~\ref{fig:playing_with_clap_agents_episode_lens} shows the results of this experiment, measured by the mean number of timesteps that it took for the user to reach the goal location while using the messages.
It shows that only the ECTL agent was able to effectively communicate with the user, with the IL agent conveying the same lack of information as the random policy.

\section{Conclusions}

In this paper, we have posed a new problem for multi-agent communication learning, namely the \textit{Cooperative Language Acquisition Problem} (CLAP).
Positioned relative to Zero-Shot Coordination and Ad Hoc Teamwork, the CLAP challenge is to build an agent that learns the communication strategy of a target community via observational data. 
But can also leverage general `environment-level skills' that transfer to any ad hoc team.
We have proposed two approaches to this problem: Imitation Learning (IL) and \textit{Emergent Communication pretraining and Translation Learning} (ECTL).

We have shown that ECTL can perform well in data scarce scenarios, including learning to communicate with a human user. Additionally, it can effectively compensate for expert demonstrations that only cover a limited distribution over the environment state space. On the other hand, while IL is brittle, it does have the potential to scale to large datasets. Further work should investigate combining the strengths of these methods.





\bibliographystyle{apalike}
\bibliography{references}

\begin{thebibliography}{}

\bibitem[Barrett et~al., 2014]{barrett_communicating_2014}
Barrett, S., Agmon, N., Hazon, N., Kraus, S., and Stone, P. (2014).
\newblock Communicating with unknown teammates.
\newblock In {\em Proceedings of the 2014 international conference on {Autonomous} agents and multi-agent systems}, {AAMAS} '14, pages 1433--1434, Richland, SC. International Foundation for Autonomous Agents and Multiagent Systems.

\bibitem[Bullard et~al., 2021]{bullard_quasi-equivalence_2021}
Bullard, K., Kiela, D., Meier, F., Pineau, J., and Foerster, J. (2021).
\newblock Quasi-{Equivalence} {Discovery} for {Zero}-{Shot} {Emergent} {Communication}.
\newblock arXiv:2103.08067 [cs].

\bibitem[Bullard et~al., 2020]{bullard_exploring_2020}
Bullard, K., Meier, F., Kiela, D., Pineau, J., and Foerster, J. (2020).
\newblock Exploring {Zero}-{Shot} {Emergent} {Communication} in {Embodied} {Multi}-{Agent} {Populations}.
\newblock arXiv:2010.15896.

\bibitem[Cope and Schoots, 2020]{cope_learning_2020}
Cope, D. and Schoots, N. (2020).
\newblock Learning to {Communicate} with {Strangers} via {Channel} {Randomisation} {Methods}.
\newblock In {\em The {Emergent} {Communication} {Workshop} at {NeurIPS}}.

\bibitem[Foerster et~al., 2016]{foerster_learning_2016}
Foerster, J., Assael, I.~A., Freitas, N.~d., and Whiteson, S. (2016).
\newblock Learning to {Communicate} with {Deep} {Multi}-{Agent} {Reinforcement} {Learning}.
\newblock In {D. D. Lee and M. Sugiyama and U. V. Luxburg and I. Guyon and R. Garnett}, editor, {\em Advances in {Neural} {Information} {Processing} {Systems} 29}, pages 2137--2145. Curran Associates, Inc.

\bibitem[Goldman and Zilberstein, 2003]{goldman_optimizing_2003}
Goldman, C.~V. and Zilberstein, S. (2003).
\newblock Optimizing information exchange in cooperative multi-agent systems.
\newblock In {\em Proceedings of the 2nd {International} {Joint} {Conference} on {Autonomous} {Agents} and {Multiagent} {Systems} ({AAMAS} 03)}, {AAMAS} '03, pages 137--144, New York, NY, USA. Association for Computing Machinery.

\bibitem[Goldman and Zilberstein, 2004]{goldman_decentralized_2004}
Goldman, C.~V. and Zilberstein, S. (2004).
\newblock Decentralized control of cooperative systems: categorization and complexity analysis.
\newblock {\em Journal of Artificial Intelligence Research}, 22(1):143--174.

\bibitem[Goldman and Zilberstein, 2008]{goldman_communication-based_2008}
Goldman, C.~V. and Zilberstein, S. (2008).
\newblock Communication-{Based} {Decomposition} {Mechanisms} for {Decentralized} {MDPs}.
\newblock {\em Journal of Artificial Intelligence Research}, 32:169--202.
\newblock arXiv:1111.0065 [cs].

\bibitem[Hu et~al., 2021]{hu_off-belief_2021}
Hu, H., Lerer, A., Cui, B., Wu, D., Pineda, L., Brown, N., and Foerster, J. (2021).
\newblock Off-{Belief} {Learning}.
\newblock In {\em the 38th {International} {Conference} on {Machine} {Learning}, {PMLR} 139}.
\newblock arXiv: 2103.04000.

\bibitem[Hu et~al., 2020]{hu_other-play_2020}
Hu, H., Lerer, A., Peysakhovich, A., and Foerster, J. (2020).
\newblock “{Other}-{Play}” for {Zero}-{Shot} {Coordination}.
\newblock In {\em Proceedings of the 37th {International} {Conference} on {Machine} {Learning}}, pages 4399--4410. PMLR.
\newblock ISSN: 2640-3498.

\bibitem[Hussein et~al., 2017]{hussein_imitation_2017}
Hussein, A., Gaber, M.~M., Elyan, E., and Jayne, C. (2017).
\newblock Imitation {Learning}: {A} {Survey} of {Learning} {Methods}.
\newblock {\em ACM Computing Surveys}, 50(2):21:1--21:35.

\bibitem[Jang et~al., 2017]{jang_categorical_2017}
Jang, E., Gu, S., and Poole, B. (2017).
\newblock Categorical {Reparameterization} with {Gumbel}-{Softmax}.
\newblock In {\em 5th {International} {Conference} on {Learning} {Representations} ({ICLR} 17)}.

\bibitem[Jaques et~al., 2019]{jaques_social_2019}
Jaques, N., Lazaridou, A., Hughes, E., Gulcehre, C., Ortega, P., Strouse, D., Leibo, J.~Z., and Freitas, N.~D. (2019).
\newblock Social {Influence} as {Intrinsic} {Motivation} for {Multi}-{Agent} {Deep} {Reinforcement} {Learning}.
\newblock In {\em Proceedings of the 36th {International} {Conference} on {Machine} {Learning}}, pages 3040--3049. PMLR.
\newblock ISSN: 2640-3498.

\bibitem[Kumar et~al., 2022]{kumar_should_2022}
Kumar, A., Hong, J., Singh, A., and Levine, S. (2022).
\newblock Should {I} {Run} {Offline} {Reinforcement} {Learning} or {Behavioral} {Cloning}?
\newblock In {\em Proceedings of the {International} {Conference} on {Learning} {Representations} ({ICLR} 22)}.

\bibitem[Lazaridou and Baroni, 2020]{lazaridou_emergent_2020}
Lazaridou, A. and Baroni, M. (2020).
\newblock Emergent {Multi}-{Agent} {Communication} in the {Deep} {Learning} {Era}.
\newblock arXiv:2006.02419 [cs].

\bibitem[Li et~al., 2023]{li_cooperative_2023}
Li, Y., Zhang, S., Sun, J., Du, Y., Wen, Y., Wang, X., and Pan, W. (2023).
\newblock Cooperative {Open}-ended {Learning} {Framework} for {Zero}-{Shot} {Coordination}.
\newblock In {\em Proceedings of the 40th {International} {Conference} on {Machine} {Learning}}, pages 20470--20484. PMLR.
\newblock ISSN: 2640-3498.

\bibitem[Lowe et~al., 2019]{lowe_pitfalls_2019}
Lowe, R., Foerster, J., Boureau, Y.-L., Pineau, J., and Dauphin, Y. (2019).
\newblock On the {Pitfalls} of {Measuring} {Emergent} {Communication}.
\newblock In {\em The 18th {International} {Conference} on {Autonomous} {Agents} and {Multiagent} {Systems} ({AAMAS})}.

\bibitem[Maddison et~al., 2017]{maddison_concrete_2017}
Maddison, C.~J., Mnih, A., and Teh, Y.~W. (2017).
\newblock The {Concrete} {Distribution}: {A} {Continuous} {Relaxation} of {Discrete} {Random} {Variables}.
\newblock In {\em 5th {International} {Conference} on {Learning} {Representations} ({ICLR} 17)}, Palais des Congrès Neptune, Toulon, France.

\bibitem[Mirsky et~al., 2020]{mirsky_penny_2020}
Mirsky, R., Macke, W., Wang, A., Yedidsion, H., and Stone, P. (2020).
\newblock A {Penny} for {Your} {Thoughts}: {The} {Value} of {Communication} in {Ad} {Hoc} {Teamwork}.
\newblock In {\em Proceedings of the {Twenty}-{Ninth} {International} {Joint} {Conference} on {Artificial} {Intelligence}}, pages 254--260. International Joint Conferences on Artificial Intelligence Organization.

\bibitem[Oliehoek and Amato, 2016]{oliehoek_concise_2016}
Oliehoek, F.~A. and Amato, C. (2016).
\newblock {\em A {Concise} {Introduction} to {Decentralized} {POMDPs}}.
\newblock Springer International Publishing, Cham.
\newblock Series Title: SpringerBriefs in Intelligent Systems.

\bibitem[Ossenkopf, 2020]{ossenkopf_comaze_2020}
Ossenkopf, M. (2020).
\newblock {CoMaze}: {A} cooperative game for zero-shot coordination.
\newblock In {\em The {Emergent} {Communication} {Workshop} at {NeurIPS}}.

\bibitem[Sarratt and Jhala, 2015]{sarratt_role_2015}
Sarratt, T. and Jhala, A. (2015).
\newblock The {Role} of {Models} and {Communication} in the {Ad} {Hoc} {Multiagent} {Team} {Decision} {Problem}.
\newblock In {\em Proceedings of the {Third} {Annual} {Conference} on {Advances} in {Cognitive} {Systems}}.

\bibitem[Stone et~al., 2010]{stone_ad_2010}
Stone, P., Kaminka, G.~A., Kraus, S., and Rosenschein, J.~S. (2010).
\newblock Ad {Hoc} {Autonomous} {Agent} {Teams}: {Collaboration} without {Pre}-{Coordination}.
\newblock In {\em Proceedings of the {Twenty}-{Fourth} {AAAI} {Conference} on {Artificial} {Intelligence}}.

\bibitem[Sukhbaatar et~al., 2016]{sukhbaatar_learning_2016}
Sukhbaatar, S., Szlam, A., and Fergus, R. (2016).
\newblock Learning multiagent communication with backpropagation.
\newblock In {\em Proceedings of the 30th {International} {Conference} on {Neural} {Information} {Processing} {Systems}}, pages 2252--2260, Barcelona, Spain. Neural Information Processing Systems.

\bibitem[Wagner et~al., 2003]{wagner_progress_2003}
Wagner, K., Reggia, J.~A., Uriagereka, J., and Wilkinson, G.~S. (2003).
\newblock Progress in the {Simulation} of {Emergent} {Communication} and {Language}.
\newblock {\em Adaptive Behavior}, 11(1):37--69.
\newblock Publisher: SAGE Publications Ltd STM.

\bibitem[Yu et~al., 2022]{yu_surprising_2022}
Yu, C., Velu, A., Vinitsky, E., Gao, J., Wang, Y., Bayen, A., and Wu, Y. (2022).
\newblock The {Surprising} {Effectiveness} of {PPO} in {Cooperative} {Multi}-{Agent} {Games}.
\newblock In {\em 36th {Conference} on {Neural} {Information} {Processing} {Systems} ({NeurIPS} 2022) {Track} on {Datasets} and {Benchmarks}.}, New Orleans, USA.

\end{thebibliography}

\appendix
\section{Appendix} \label{app:training_params}

\subsection{Environment Details}

\textbf{Goal Communications Gridworld.}
Each agent has a goal square in the grid that they need to reach.
At the start of each episode, this goal is sampled from a uniform distribution over the squares.
There are five actions: move up, down, right, left, or stay put.
Agents do not physically interact on the grid, and so can be in the same positions.
For each agent, if they have not reached their goal then the team is deducted -1.
When they reach the goal, a reward of +1 is given, and if they have already reached it in the past then no penalty or reward is given.
An episode ends once all agents have reached their goals or a time limit is reached.
Each agent's observation is the concatenation of the following vectors:
\begin{itemize}[noitemsep]
    \item A binary vector of nine numbers, indicating if the goal is in one of the nine squares near the agent.
    \item One-hot encodings of the $(\hat{g}_x, \hat{g}_y)$ coordinates corresponding to a `close guess' of another agent's goal. This close guess is sampled at the start of an episode from a uniform distribution of locations within one tile away from the true goal.
    \item One-hot encodings of each of the $(p_x, p_y)$ position coordinates for each agent in the environment.
\end{itemize}

\begin{table*}
    \begin{subtable}{.4\textwidth}
        \centering
        \caption{MAPPO hyperparameters for training teams on the gridworld problem.}
        \label{tab:mappo_hyperparams_grid}
        \begin{tabular}{ll}
            \toprule
            Hyperparameter & Value \\
            \midrule
            Message dimension $(|\Sigma|)$ & $5$ \\
            Optimiser & Adam \\
            Learning rate & $0.0005$ \\
            Adam Betas & (0.9, 0.999) \\
            Gumbel-Softmax Temperature & $2.0$ (no annealing) \\
            SGD Minibatch Size & $2,048$ \\
            Training Batch Size & $10,000$ \\
            SGD Num Iterations & $5$ \\
            Value Function Clip Parameter & $10$ \\
            Value Function Loss Coef & $0.25$ \\
            Lambda (GAE/PPO) & $0.95$ \\
            KL Coef (PPO) & 0 \\
            $enc$ Hidden Layer Sizes & [256, 256] \\
            $\pi^e$ Hidden Layer Sizes & [256] \\
            $vf$ Hidden Layer Sizes & [256, 256] \\
            \bottomrule
        \end{tabular}
    \end{subtable}%
    \hfill
    \begin{subtable}{.45\textwidth}
        \centering
        \caption{MAPPO hyperparameters for training teams on the driving problem.}
        \label{tab:mappo_hyperparams_driving}
        \begin{tabular}{ll}
            \toprule
            Hyperparameter & Value \\
            \midrule
            Message dimension $(|\Sigma|)$ & $16$ \\
            Optimiser & Adam \\
            Learning rate & $0.0005$ \\
            Adam Betas & (0.9, 0.999) \\
            Gumbel-Softmax Temperature & $5.0$ (no annealing) \\
            SGD Minibatch Size & $2,048$ \\
            Training Batch Size & $200,000$ \\
            SGD Num Iterations & $10$ \\
            Value Function Clip Parameter & $10$ \\
            Value Function Loss Coef & $0.25$ \\
            Lambda (GAE/PPO) & $0.95$ \\
            KL Coef (PPO) & 0 \\
            $enc$ Hidden Layer Sizes & [256, 256] \\
            $\pi^e$ Hidden Layer Sizes & [256] \\
            $vf$ Hidden Layer Sizes & [256, 256] \\
            \bottomrule
        \end{tabular}
    \end{subtable}
\end{table*}

For all experiments we keep the grid size fixed at $(5\times 5)$ and the maximum number of timesteps at 10.

\textbf{Goal Communications Driving Game.} In this game, each agent steers and accelerates a body in a continuous grid (Figure \ref{fig:pit_env_illustration}).
The agents must navigate to one of eight fixed goal locations, known by their partner and selected at random for each episode.
Each agent observes the positions, previous positions, angles, and velocities of all the agents, represented as vectors $\mathds{R}^2$.
Each agent observes the goal location of another agent, giving rise to the need to communicate.

There are four actions that an agent can take: (1) turn clockwise by $\theta$ radians, (2) turn anti-clockwise by $\theta$ radians, (3) accelerate by $a$ units per second/second.
The value $\theta$ is fixed at $\pi/8$ and $a$ at 1.0.
Over each timestep, 0.2 seconds of simulated time elapse, i.e. the game runs in `real-time' at 5 frames per second.
The team has a maximum of 200 timesteps to complete the task, or the episode ends when all the goals have been reached.
There are two scenarios for this environment defined by the presence or absence of a circular region called the `pit' in the centre of the world where a large penalty is applied for every agent within the region.

Each agent's reward at each time step is computed from four components:
\begin{enumerate}
    \item $c_{pit}$ is the penalty for the agent being in the pit location, if the pit scenario is selected.
    \item $\alpha \Delta d_{min}$, where $\Delta d_{min}$ is the change in closest distance to the goal reached during the episode. $\alpha$ is a hyperparameter set to 200. This term is zero if the distance to the goal is changing, but it is also not given if the agent is in the pit. However, the `closest distance to the goal' is still updated while in the pit, so the agent is penalised directly and indirectly through the lost opportunity of being awarded the bonus.
    \item $c_{time}$ is a penalty given at each time step to encourage completing the task quickly. Set to a constant of 0.1.
    \item $r_{goal}$ is a bonus given for reaching the goal. It is set to 500.0 and only awarded once at the timestep that the goal is reached.
\end{enumerate}
After an agent has already reached its goal, it is always awarded zero reward unless it is in the pit.

\subsection{Training Hyperparameters}

All reinforcement learning was done using RLlib and Pytorch.
The hyperparameters for MAPPO training in each environment can be found in Tables \ref{tab:mappo_hyperparams_grid} and \ref{tab:mappo_hyperparams_driving}.
Note the key differences being the temperature of the Gumbel-Softmax communication channel, which was increased to improve learning, and the size of the training batch that needed to dramatically increase.
IL and ECTL just use Pytorch for training and RLlib for evaluation.
For IL both components of the $\pi_{im}$ are feed-forward networks with hidden layer sizes [256].
Likewise, for ECTL, both translation functions have hidden layer sizes [256, 256].
The full hyperparameters for training with the different methods are given in Table~\ref{tab:joiner_training_params}.

\begin{table}[t]
\centering
\caption{Signalling/Listening training hyperparameters (for both IL and ECTL).}
\label{tab:joiner_training_params}
\begin{tabular}{ll}
\toprule
Hyperparameter & Value \\
\midrule
Optimiser & SGD with momentum \\
Learning rate & $0.001$ \\
Momentum & 0.9 \\
Minibatch Size & $1024$ \\
Weight Decay & $1\times 10^{-5}$ \\
Epochs & 1500 \\
\bottomrule
\end{tabular}
\end{table}

\subsection{Evaluation Methodology}

For evaluation on the CLAP-Replace tasks, we took each trained community and used each permutation of teams to create joiner/target community pairs. We trained joiner agents using IL and ECTL for each possible replaceable agent (2 agents in driving and 3 agents in gridworld), and then for each configuration we conducted 3 trials and performed evaluations. Each configuration runs the entire CLAP-Replace and training pipeline from scratch (i.e. independent data collection and random seeds). Each trained joiner agent was then evaluated for 500 episodes in the zero-shot CLAP-Replace setting.
For Figure~\ref{fig:reward_vs_collect_eps}, for each sample along the $x$-axis (different $N_{collect}$ values), this entire process was run. Thus there was 9 $N_{collect}$ samples $\times$ 2 community permutations $\times$ 2 possible replacement agents $\times$ 3 trials, totally 432 training runs.

\begin{figure*}
    \centering
    \includegraphics[width=0.7\textwidth]{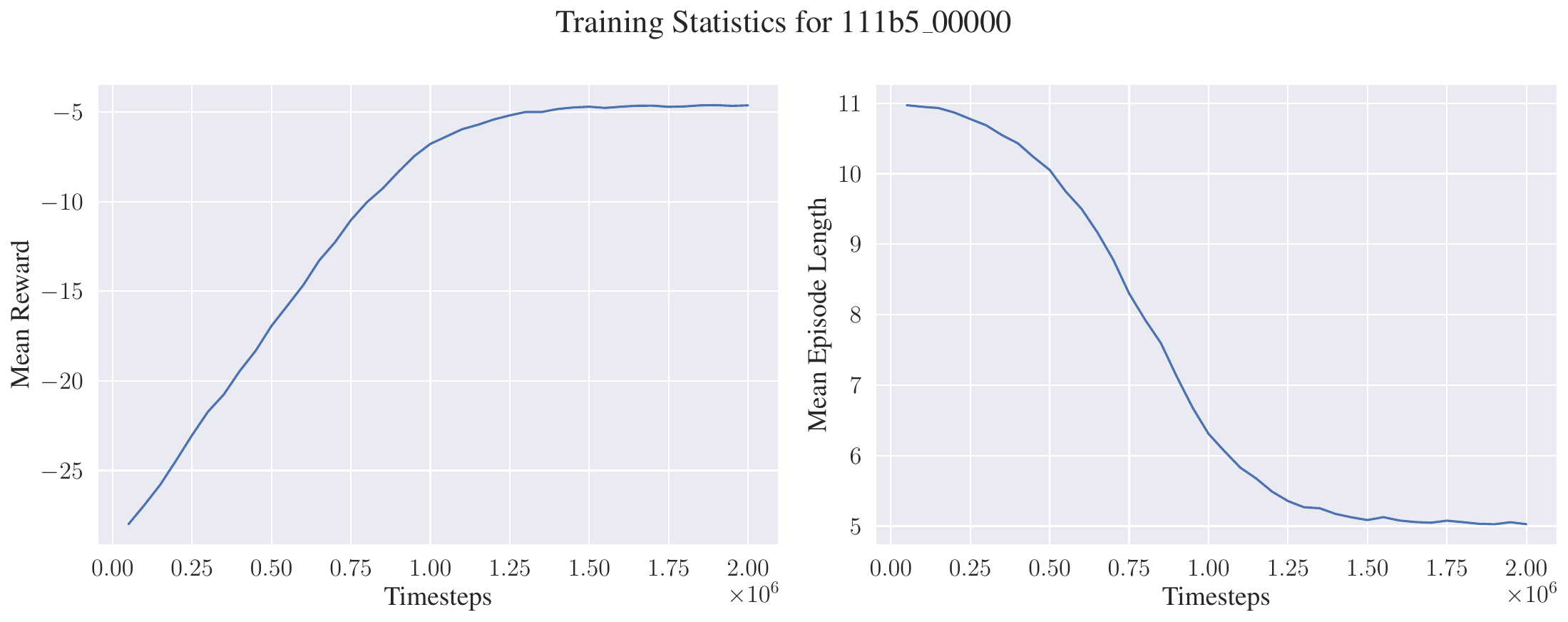}
    \caption{Training curves for one training run on the gridworld environment.}
    \label{fig:mappo-learning-curves-gridworld-1}
\end{figure*}
\begin{figure*}
    \centering
    \includegraphics[width=0.7\textwidth]{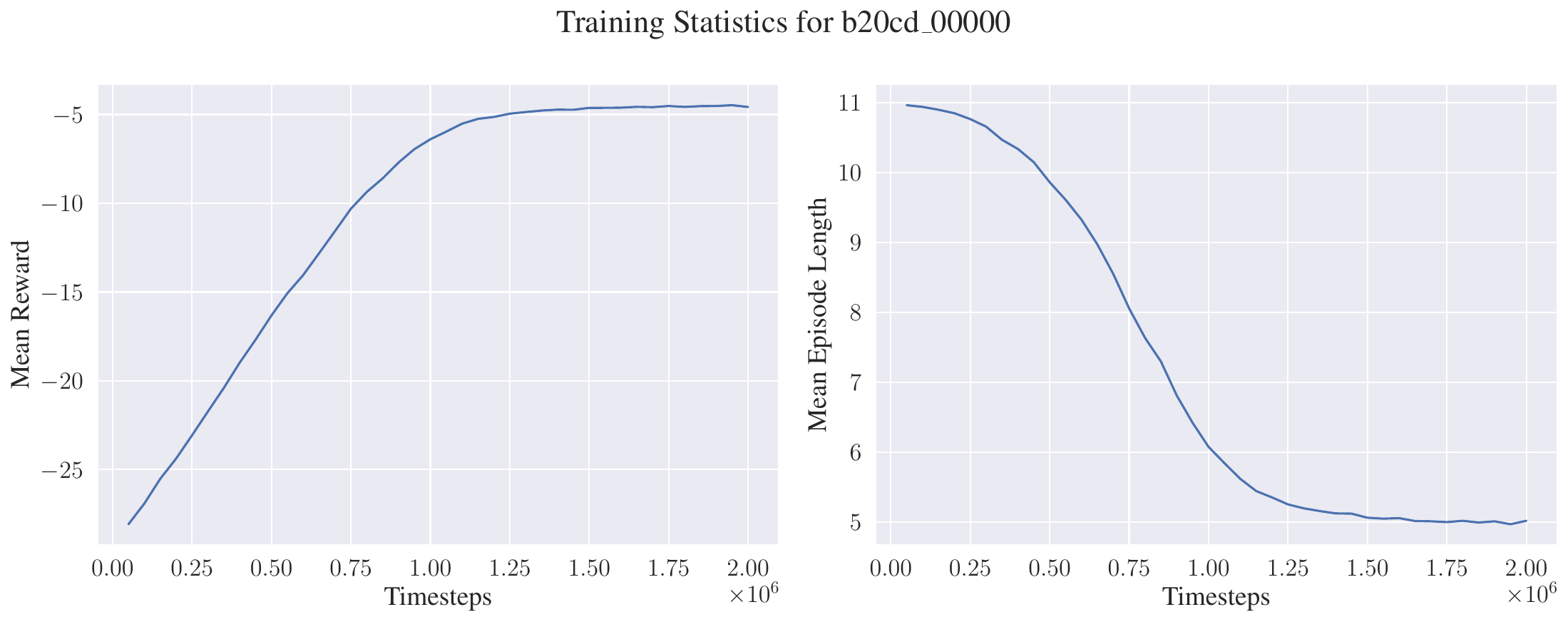}
    \caption{Training curves for one training run on the gridworld environment.}
    \label{fig:mappo-learning-curves-gridworld-1}
\end{figure*}
\begin{figure*}
    \centering
    \includegraphics[width=0.7\textwidth]{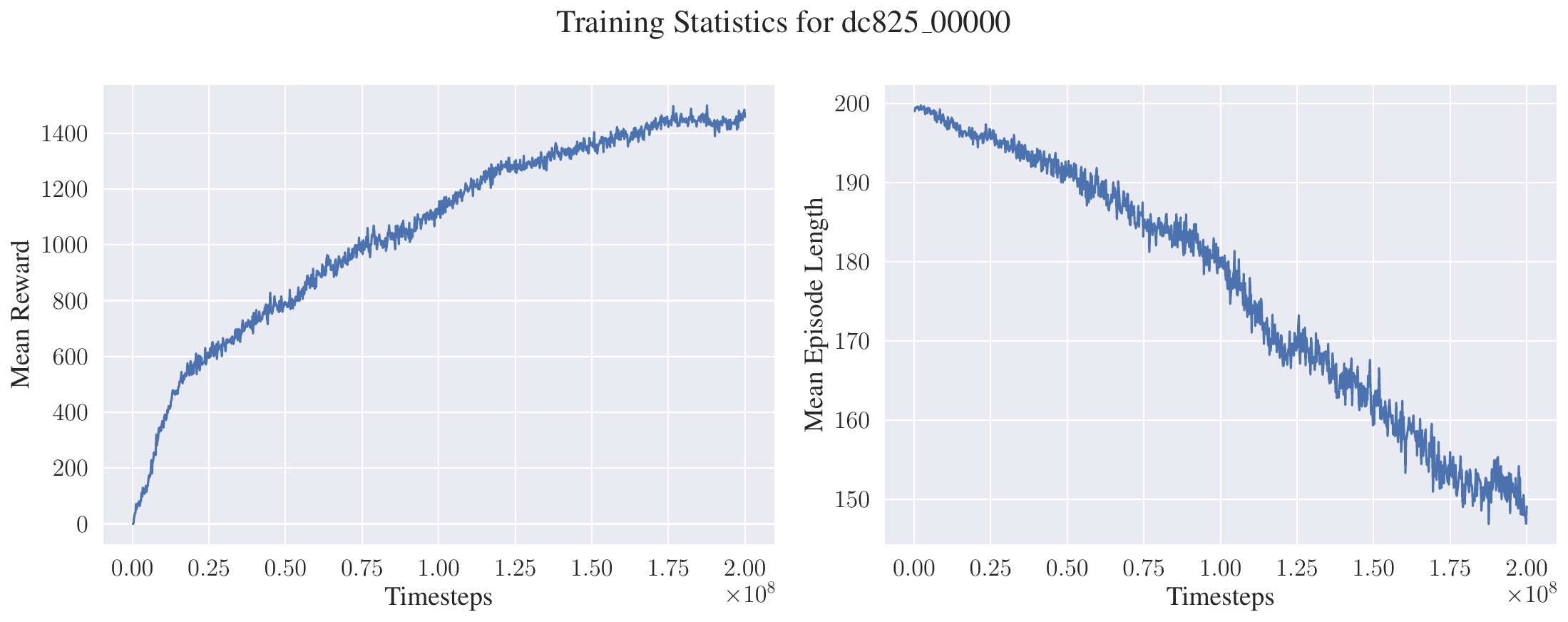}
    \caption{Training curves for one training run on the driving environment, in the `pit' scenario.}
    \label{fig:mappo-learning-curves-gridworld-1}
\end{figure*}
\begin{figure*}
    \centering
    \includegraphics[width=0.7\textwidth]{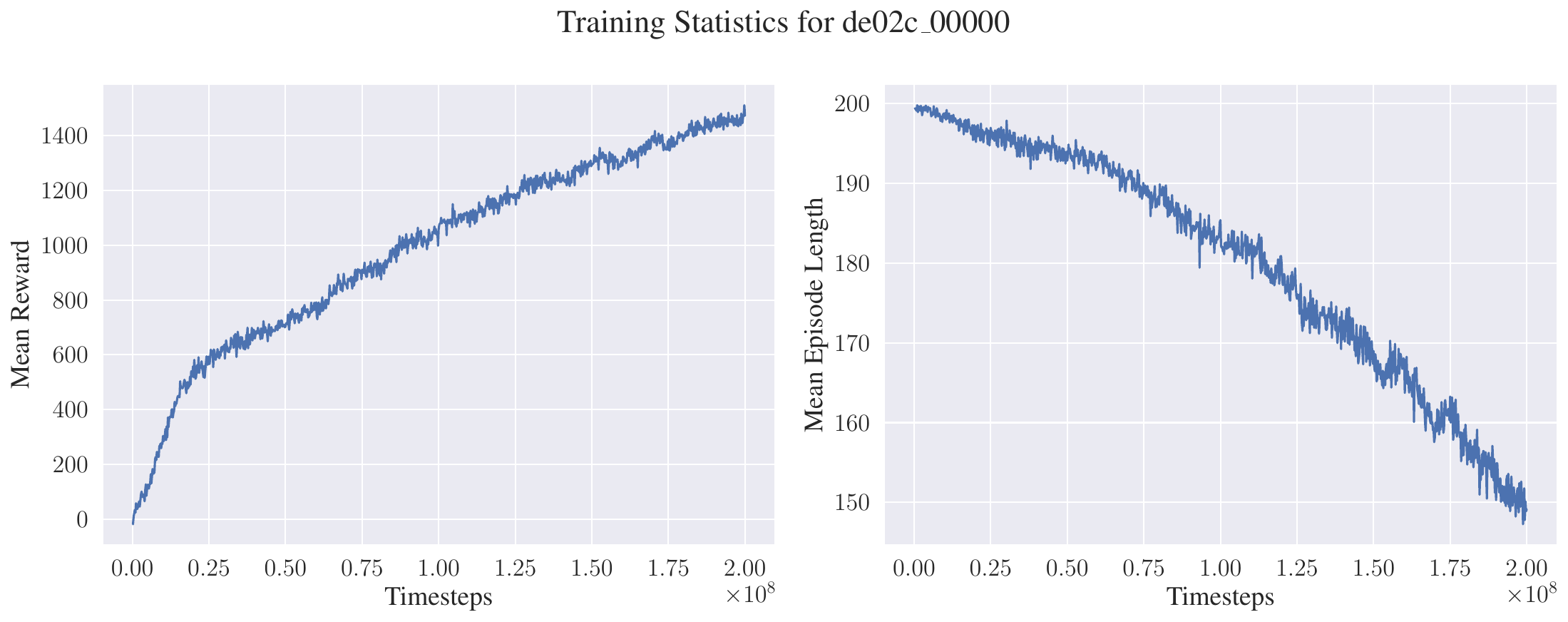}
    \caption{Training curves for one training run on the driving environment, in the `pit' scenario.}
    \label{fig:mappo-learning-curves-gridworld-1}
\end{figure*}
\begin{figure*}
    \centering
    \includegraphics[width=0.7\textwidth]{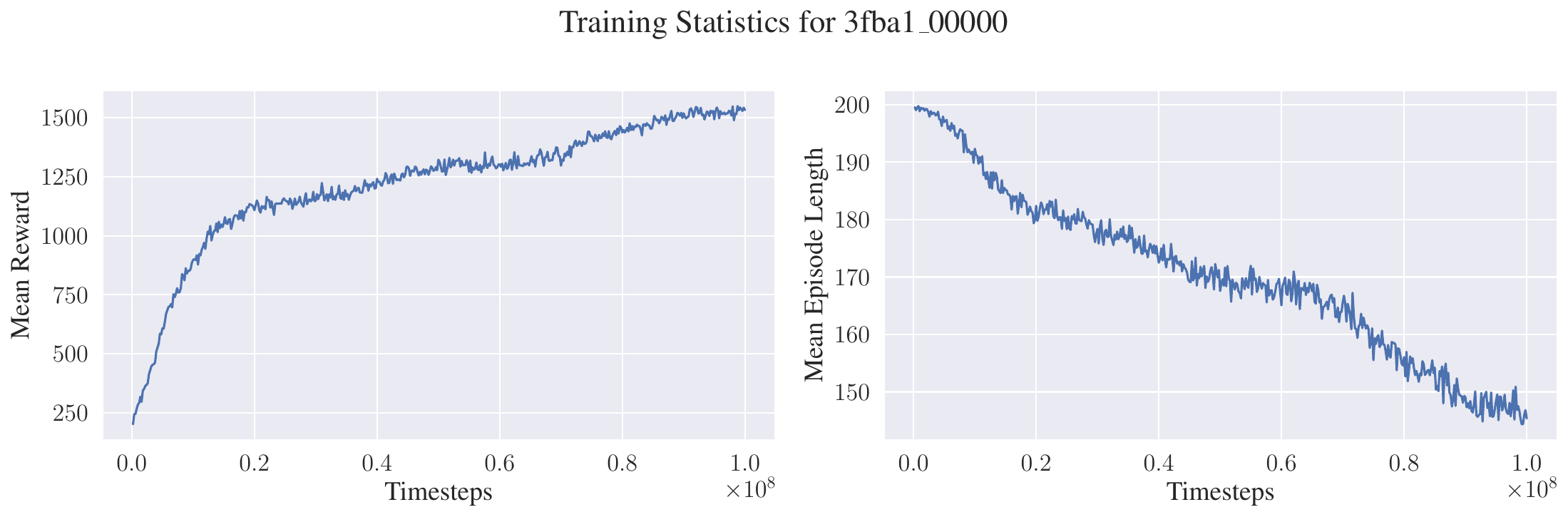}
    \caption{Training curves for one training run on the driving environment, in the `no pit' scenario.}
    \label{fig:mappo-learning-curves-gridworld-1}
\end{figure*}
\begin{figure*}
    \centering
    \includegraphics[width=0.7\textwidth]{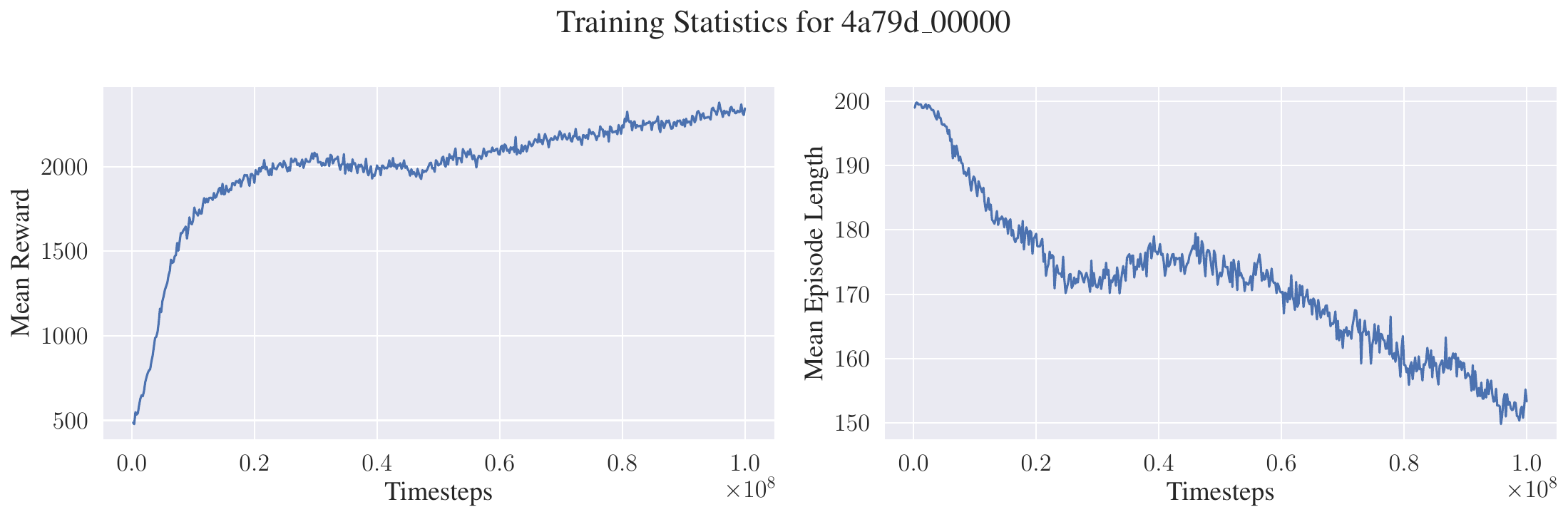}
    \caption{Training curves for one training run on the driving environment, in the `no pit' scenario.}
    \label{fig:mappo-learning-curves-gridworld-1}
\end{figure*}
\end{document}